\documentclass{article} 
\usepackage{iclr2025_conference,times}

\iclrfinalcopy  

\usepackage{amsmath,amsfonts,bm}

\def\eqref#1{equation~\ref{#1}}

\def\1{\bm{1}}

\DeclareMathAlphabet{\mathsfit}{\encodingdefault}{\sfdefault}{m}{sl}
\SetMathAlphabet{\mathsfit}{bold}{\encodingdefault}{\sfdefault}{bx}{n}

\usepackage{hyperref}
\usepackage{url}

\usepackage{graphicx}

\usepackage{algorithm}
\usepackage{algpseudocode}
\algnewcommand\algorithmicforeach{\textbf{for each}}
\algdef{S}[FOR]{ForEach}[1]{\algorithmicforeach\ #1\ \algorithmicdo}
\usepackage[skins]{tcolorbox}
\usepackage{soul}
\usepackage{subcaption}
\usepackage{mathabx}

\title{MATEY: multiscale adaptive foundation \\ models for spatiotemporal physical systems}

\author{
  Pei Zhang$^{1}$, M.~Paul Laiu$^{2}$, Matthew Norman$^{3}$, Doug Stefanski$^{1}$, John Gounley$^{1}$ 
  \\
  $^{1}$Computational Sciences and Engineering Division \\
  $^{2}$Computer Science and Mathematics Division\\
  $^{3}$National Center for Computational Science\\
  Oak Ridge National Laboratory\\
  1 Bethel Valley Road, Oak Ridge, TN 37830, USA \\
\texttt{\{zhangp1, laiump, normanmr, stefanskidl, gounleyjp\}@ornl.gov} \\
}

\begin{document}

\maketitle

\begin{abstract}

Accurate representation of the multiscale features in spatiotemporal physical systems using vision transformer (ViT) architectures requires extremely long, computationally prohibitive token sequences. To address this issue, we propose two adaptive tokenization schemes that dynamically adjust patch sizes based on local features: one ensures convergent behavior to uniform patch refinement, while the other offers better computational efficiency. 
Moreover, we present a set of spatiotemporal attention schemes, where the temporal or axial spatial dimensions are decoupled, and evaluate their computational and data efficiencies.
We assess the performance of the proposed multiscale adaptive model, MATEY, in a sequence of experiments. 
The results show that adaptive tokenization schemes achieve improved accuracy without significantly increasing the length of the token sequence. 
Compared to a full spatiotemporal attention scheme or a scheme that decouples only the temporal dimension, we find that fully decoupled axial attention is less efficient and expressive, requiring more training time and model weights to achieve the same accuracy. 
Finally, we demonstrate in two fine-tuning tasks featuring different physics that models pretrained on PDEBench data outperform the ones trained from scratch, especially in the low data regime with frozen attention.

\end{abstract}

\section{Introduction}\label{sec-intro}

Developing foundation models for physical systems is vital for energy generation, earth sciences, and power and propulsion systems. These models offer faster solutions than physics-based simulations and can generalize better across multiple systems than single-purpose AI approaches. However, their application to physical systems, often characterized by multiple sub-processes at different scales, is still in the early stages. For instance, fluid flowing around a cylinder creates a von Kármán vortex street, a highly dynamic flow with rapidly evolving vortices. Accurate solutions of such multiscale systems require a very high resolution representation to capture the most complex features across space and time. However, for scientific machine learning as for modeling and simulation, using very high resolutions to achieve accurate solutions incurs significant computational cost. This is particularly true for developing foundation models using vision transformer (ViT)-based architectures, as using the standard self-attention mechanism for extremely long spatiotemporal sequences can become prohibitively computationally expensive.

Efficient representation of multiscale features in high-resolution inputs has been an active research topic in computer vision. Three broad approaches can be characterized. First, multiscale models like Swin Transformer \citep{liu2021swin} and MViTv2 \citep{li2022mvitv2} introduce multiple stages with decreasing resolution and increasing feature dimension for efficient hierarchical representations.
Second, computational techniques have been developed that facilitate training on long sequences (e.g., sequence parallelism across GPUs \citep{jacobs2023deepspeedulyssesoptimizationsenabling}) or reduce the effective sequence length in the attention kernel (e.g., decomposing attention along axial directions \citep{ho2019axial}). Third, the actual sequence length can be directly shortened by pruning and merging tokens (\citep{Haurum_2023_ICCV, meng2022adavit, yin2022advit, Bolya_2023_CVPR}), though this strategy may lead to critical information loss \citep{liu2024sorareviewbackgroundtechnology}.

These techniques have recently been adopted in scientific machine learning (sciML) for physical systems. For example, the atmosphere foundation model Aurora \citep{bodnar2024aurora} uses Swin Transformer, while axial attention is applied by MPP \citep{mccabe2023multiple}. Despite the progress, computational constraints remain a bottleneck, as existing approaches do not yet handle high-fidelity solutions of applications such as computational fluid dynamics, in which input sequences can easily exceed billions of tokens. More efficient algorithms are needed to enable the development of foundation models for multiscale multiphysics systems.

In this work, we develop a multiscale adaptive foundation model, MATEY (see Figure~\ref{fig-matey}), that provides two key algorithmic contributions to address the challenges posed by spatiotemporal physical systems. First, inspired by the adaptive mesh refinement (AMR) technique, we introduce an adaptive tokenization method that dynamically adjusts patch sizes across the system based on local features, which provides as much as a $2\times$ reduction in compute for similar or higher accuracy. Second, we present a set of spatiotemporal attention schemes based on the axial attention \citep{ho2019axial} that differ in their decomposition of long spatiotemporal sequences and identify the cost in time-to-accuracy for axial attention. Finally, we assess the fine-tuning performance of models pretrained on PDEBench \citep{takamoto2022pdebench} in two highly out-of-distribution settings, colliding thermals and magnetohydrodynamics (MHD), that include additional physical variables not included in pretraining and observe the pretrained models outperforming random initialized models.

\section{Related work}\label{sec-related}

\paragraph{Scientific foundation models.} 
Several research directions have been explored for building foundation models for physical systems, including multiple physics pretraining \citep{mccabe2023multiple} with PDEBench data, input augmentation with PDE system configurations \citep{hang2024unisolver}, robust pretraining schemes \citep{hao2024dpot}, fine-tuning effectiveness investigations \citep{subramanian2024towards}, and data-efficient multiscale ViT architectures \citep{herde2024poseidon}.
While these work made remarkable progress, they do not directly address the issue of token sequence length, which becomes a computation bottleneck when applying ViTs to high dimension or high resolution data.

\paragraph{Multiscale ViTs.}
While most multiscale ViTs achieve hierarchical representations via multi-stage attention blocks at different resolutions (e.g., MViTv2 \citep{li2022mvitv2} and Swin Transformer \citep{liu2021swin}), there are a few focusing on tokenization schemes, such as\citep{yin2022advit,fan2024vitarvisiontransformerresolution,zhang2024adaptivepatchinghighresolutionimage,havtorn2023msvit}.
Among these, the single-stage MSViT with dynamic mixed-scale tokenization \citep{havtorn2023msvit}, which leverages a learnable gating neural network for selecting the token refinement,
is most related to our work. 
This approach requires a tailored gate loss function and an adaptive trimming scheme to handle the high overhead cost, which in return hurts gate training accuracy. 
In contrast, the tokenization scheme in MATEY adaptively adjusts the patch sizes directly based on local feature scales, which is simpler and more direct. 

\paragraph{Axial attentions.} 
The quadratic scaling nature of attention makes it computationally prohibitive for extremely long token sequences from multidimensional systems. To address this challenge, \citep{ho2019axial} proposed the axial attention, which decomposes the full attention into a sequence of attention operations along each axis. It reduces the attention cost from $\mathcal{O}(N^{2d})$ to $\mathcal{O}(N^{d+1})$, for a given $d$-dimensional system with $N^d$ tokens. ViViT \citep{arnab2021vivitvideovisiontransformer} factorized the spatiotemporal attention into spatial- and temporal-dimensions for video classification. \citep{mccabe2023multiple} applied the axial attention in the Axial ViT (AViT) for spatiotemporal solutions of physical systems. While these spatiotemporal attention schemes can reduce the sequence length and hence the attention cost, their impact on accuracy in physical systems is unclear.

\section{MATEY, explained}\label{sec-method}

\begin{figure}[h]
\centering
\includegraphics[width=0.7\textwidth]{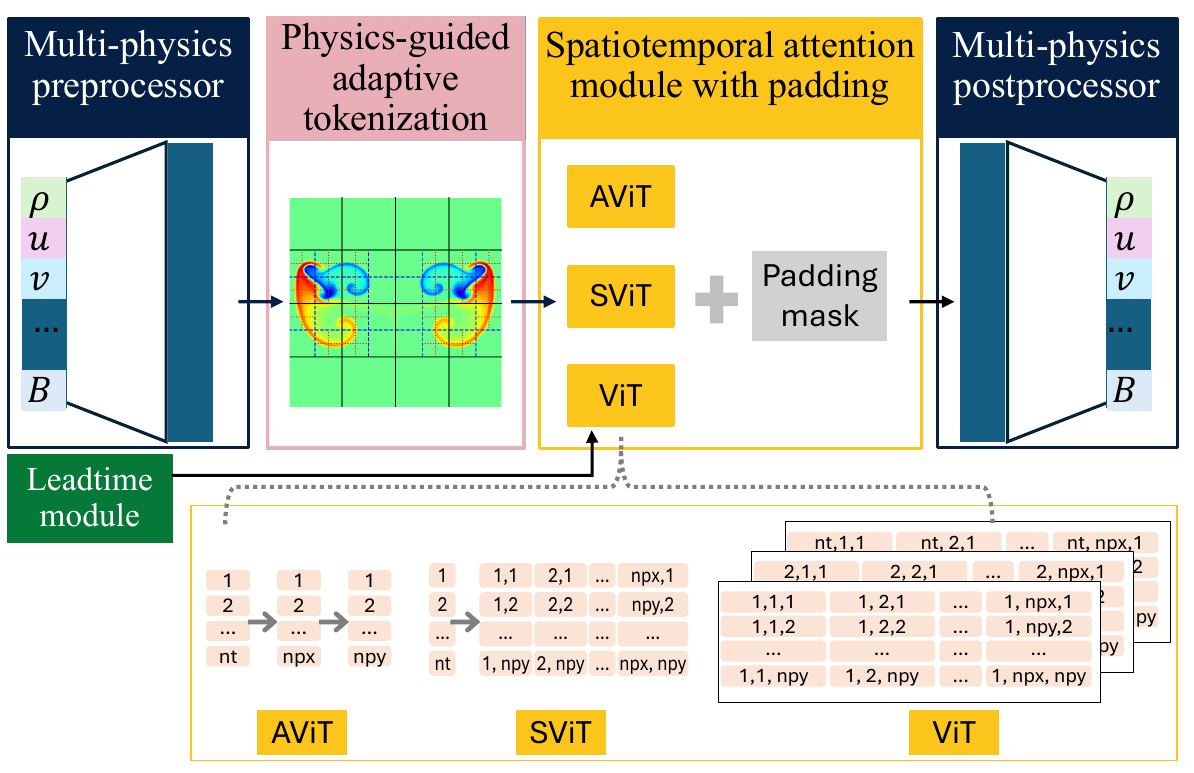}
\includegraphics[width=0.8\textwidth]{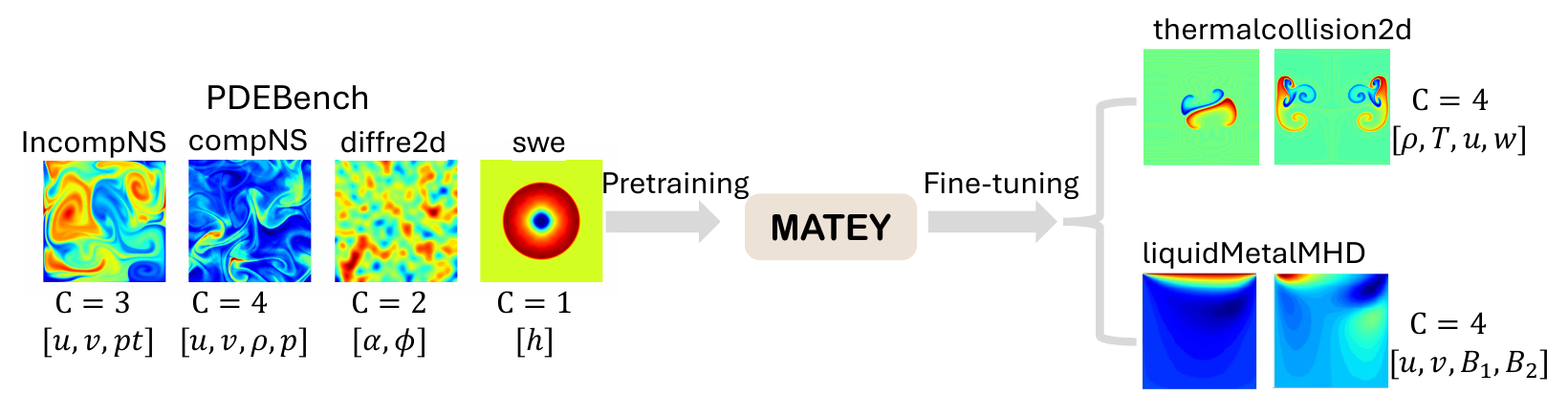}
\caption{MATEY: multiscale adaptive foundation models for spatiotemporal physical systems.}
\label{fig-matey}
\end{figure}

We propose multiscale adaptive foundation models, MATEY, to predict two-dimensional spatiotemporal solutions of multiple physical systems. The architecture of MATEY is illustrated in Figure~\ref{fig-matey}. 
Given a sequence of $T$ past solutions of some physical system leading up to time $t$, MATEY predicts the solution at a future time $t+t_{\text{lead}}$ by learning from sequences of solutions for multiple physical systems. 
Specifically, MATEY learns a model $\mathbf{f}_{\mathbf{w}}$ such that $\bold u_{t+t_{\text{lead}}} \approx \bold f_\bold w(\bold u_{t-T+1},\ldots,\bold u_{t}; t_{\text{lead}})$ by training parameters $\mathbf{w}$ to minimize the loss of the prediction from the solution sequence $\bold{U}=[\bold u_{t-T+1},\ldots,\bold u_t]$ against the future solution with a lead time $\bold u_{t+t_{\text{lead}}}$. 
In the following paragraphs, we give detailed descriptions for each component in MATEY.

\paragraph{Multi-physics preprocessor, postprocessor, and training.} 
To accommodate multiple physical systems with different sets of variables at different spatial resolutions, we adopt the multi-physics preprocessor and postprocessor used in MPP \citep{mccabe2023multiple}. 
For system $k$ with $C_k$ variables, the preprocessor first encodes solutions $\mathbf{u}_t(x,y)\in\mathbb{R}^{C_k}$ to a latent space $\mathbb{R}^{C_{\textup{uni}}}$, where $C_{\textup{uni}}\gg C_k$ is shared among all systems. 
Specifically, letting $H$ and $W$ denote the resolution in the $x$ and $y$ directions, respectively, the preprocessor encodes the solution $\bold{U}_k\in \mathbb{R}^{T\times H\times W \times C_k}$ of system $k$ into the unified latent representation $\bold{U}\in \mathbb{R}^{T\times H\times W \times C_\textup{uni}}$.
$\bold{U}$ is then tokenized into sequences $\boldsymbol{Z}^0\in \mathbb{R}^{nt\times npx\times npy \times C_\textup{emb}}$ in the tokenization module, which consists of convolutional blocks.
Here $nt=T/p_t$, $npx=H/p_x$, and $npy=W/p_y$ are the number of patches in each dimension with prescribed patch size $[p_t, p_x, p_y]$.
After passing through $L$ attention blocks, the input token sequence $\boldsymbol{Z}^0$ leads to the attention output $\boldsymbol{Z}^L\in \mathbb{R}^{nt\times npx\times npy \times C_\textup{emb}}$. 
The last temporal snapshot of $\boldsymbol{Z}^L$ is then decoded in the postprocessor into the prediction $\mathbf{u}_{\textup{pred}}\in \mathbb{R}^{H\times W \times C_k}$.
In this work, the preprocessor is a linear map, the tokenization module is implemented as a convolutional neural network (CNN), and the final decoding postprocessor uses 2D transposed convolutional blocks.
To train the model from solutions with different resolutions, we follow the approach in MPP by performing system-based sampling in the training process and fusing information from samples across different systems via multi-GPU training with PyTorch Distributed Data Parallelism (DDP) and gradient accumulation.

\paragraph{Attention mechanisms --- AViT, SViT, and ViT.}
The standard ViT attention mechanism takes into account the attention across the entire set of spatiotemporal dimensions, which results in a high attention cost when extremely long spatiotemporal token sequences (e.g., from high-resolution spatiotemporal data) are considered.
To address this issue, various factorized attention mechanisms have been proposed, such as AViT \citep{ho2019axial, mccabe2023multiple} and a spatio-temporal decoupled attention \citep{arnab2021vivitvideovisiontransformer}, referred to as SViT here.  
These attention mechanisms mainly consist of the same multihead self-attention (MHSA) and feed forward multi-layer perceptron (MLP) but differ in their attention block architecture. When $L$ attention blocks are cascaded, the standard attention block in ViT is given as
\begin{equation}\label{eq:attention}
\begin{aligned} 
    \widehat{\boldsymbol{Z}}^{0} &= \boldsymbol{Z}^{0}+ \boldsymbol{E}_\textup{pos},
    \quad 
     \boldsymbol{Z}^{0} = [\boldsymbol{z}^0_1, \boldsymbol{z}^0_{2},\ldots, \boldsymbol{z}^0_{N}], \\
    \boldsymbol{Z}^{1} &= \text{MLP}(\widetilde{\boldsymbol{Z}}^{1})+ \widetilde{\boldsymbol{Z}}^{1} ,  \qquad     
    \widetilde{\boldsymbol{Z}}^{1} = \text{MHSA} (\widehat{\boldsymbol{Z}}^{0})+ \widehat{\boldsymbol{Z}}^{0}+ \text{MLP}(t_{\textup{lead}}),\\
    \boldsymbol{Z}^{\ell} &= \text{MLP}(\widetilde{\boldsymbol{Z}}^{\ell})+ \widetilde{\boldsymbol{Z}}^{\ell},  \qquad \widetilde{\boldsymbol{Z}}^{\ell} = \text{MHSA} ({\boldsymbol{Z}}^{\ell-1})+ {\boldsymbol{Z}}^{\ell-1}, \qquad \ell=2, \ldots, L   
\end{aligned}
\end{equation}
where $[\boldsymbol{z}^0_1,\ldots, \boldsymbol{z}^0_{N}]$ denotes the full spatiotemporal token sequence of length $N$ with each token $\boldsymbol{z}^0_i\in\mathbb{R}^{C_{\textup{emb}}}$, $\boldsymbol{E}_\textup{pos}$ is a positional embedding term, and each MHSA and MLP is followed by an \texttt{InstanceNorm1d} module.
In ViT, the token sequence includes all spatiotemporal patches, meaning $N=nt\cdot npx \cdot npy$,
resulting in an overwhelming cost of $\mathcal{O}((nt\cdot npx \cdot npy)^2)$ operations for attention.
In contrast, SViT decouples the attention into $npx\cdot npy$ time-attention blocks and $nt$ space-attention blocks cascaded sequentially, as in ``$\textup{MHSA}_\textup{time} \rightarrow \textup{MHSA}_\textup{space}\rightarrow \textup{MLP}$", 
 \begin{equation}\label{eq:svit-att}
 \begin{aligned}
&\textup{Time sequences:}\quad {\boldsymbol{Z}}^{\ell-1}_i=\left[\boldsymbol{z}^{\ell-1}_{(i-1)\cdot nt+1},\boldsymbol{z}^{\ell-1}_{(i-1)\cdot nt+2},\ldots, \boldsymbol{z}^{\ell-1}_{(i-1)\cdot nt+nt}\right], i =1, \ldots, npx\cdot npy\\
&\textup{Attention in time:}\quad  \boldsymbol{Z}^{\ell-\frac{1}{2}}_i = \text{MHSA}_\textup{time} \left({\boldsymbol{Z}}^{\ell-1}_i\right)+ {\boldsymbol{Z}}^{\ell-1}_i, \quad i =1, \ldots, npx\cdot npy\\
&\textup{Space sequences:}\quad \widecheck{\boldsymbol{Z}}^{\ell-\frac{1}{2}}_t=\left[\boldsymbol{z}^{\ell-\frac{1}{2}}_t,\boldsymbol{z}^{\ell-\frac{1}{2}}_{t+nt},\ldots, \boldsymbol{z}^{\ell-\frac{1}{2}}_{t+nt \cdot (npx\cdot npy -1)}\right], \quad t=1, \ldots, nt,\\
&\textup{Attention in space:}\quad \widetilde{\boldsymbol{Z}}^{\ell}_t = \text{MHSA}_\textup{space} \left({\widecheck{\boldsymbol{Z}}}^{\ell-\frac{1}{2}}_t\right)+ {\widecheck{\boldsymbol{Z}}}^{\ell-\frac{1}{2}}_t, \quad t=1, \ldots, nt,\\
&\textup{Feed forward ML:}\quad    \boldsymbol{Z}^{\ell}= \text{MLP}\left(\widetilde{\boldsymbol{Z}}^{\ell}\right)+ \widetilde{\boldsymbol{Z}}^{\ell},\quad \ell=1, \ldots, L,
\end{aligned}
\end{equation}
which reduces the MHSA cost to $npx\cdot npy\cdot\mathcal{O}(nt^2)+ nt\cdot\mathcal{O}((npx \cdot npy)^2)$. The position embedding and the lead time MLP are omitted in (\ref{eq:svit-att}) for simplicity.
AViT further decomposes the space-attention in SViT into two axial directions following the same approach, which leads to a cost of $npx \cdot npy \cdot\mathcal{O}(nt^2) + nt \cdot npy \cdot \mathcal{O}(npx^2) + nt\cdot npx\cdot\mathcal{O}(npy^2)$.
The decomposition in both AViT and SViT neglects some spatiotemporal correlations, and thus gives shorter token sequence length for each attention block, at the cost of introducing additional attention blocks. These extra attention blocks moderately increase the model size, as shown in Table~\ref{tab-modelsize}. Note that within the same size category considered in Table~\ref{tab-modelsize}, AViT and ViT are larger than ViT due to the additional $\textup{MHSA}$, while AViT and ViT have similar sizes because AViT reuses the same attention blocks for different spatial directions.
In MATEY, we implement the three attention mechanisms -- AViT, SViT, and ViT -- and evaluate their performance on test problems to study how the lost spatiotemporal correlations affect the quality of the solution and to assess the impact of decoupled attentions with additional attention blocks on the learning efficiency for multi-physics foundation models.

 \paragraph{Adaptive tokenization.}
 
Smaller patch sizes are preferred for better representation accuracy, as ViTs can capture long-range correlations between patches well but lack inductive biases within patches. However, features in physical systems often cross multiple length scales and exhibit strong spatiotemporal inhomogeneities. Consequently, constant patch sizes that are small enough to provide good accuracy in the necessary regions of such systems result in impractically long token sequence lengths over the entire domain. To address this issue, we propose an adaptive ViT that dynamically adjusts the tokenization patch sizes according to local physical features. To maximize expressiveness, we start with coarse patching and identify the most complex patches in each sample based on a simple metric, such as the variance of local features. The identified patches are further refined to the sub-token-scale (STS) to improve representation accuracy. Adaptive patch size leads to patches of varying length across samples, which are handled with padding masks. Patch position and patch area bias are represented following the embedding method in \citep{bodnar2024aurora}.

\begin{figure}[h]
\centering
\includegraphics[width=0.8\textwidth]{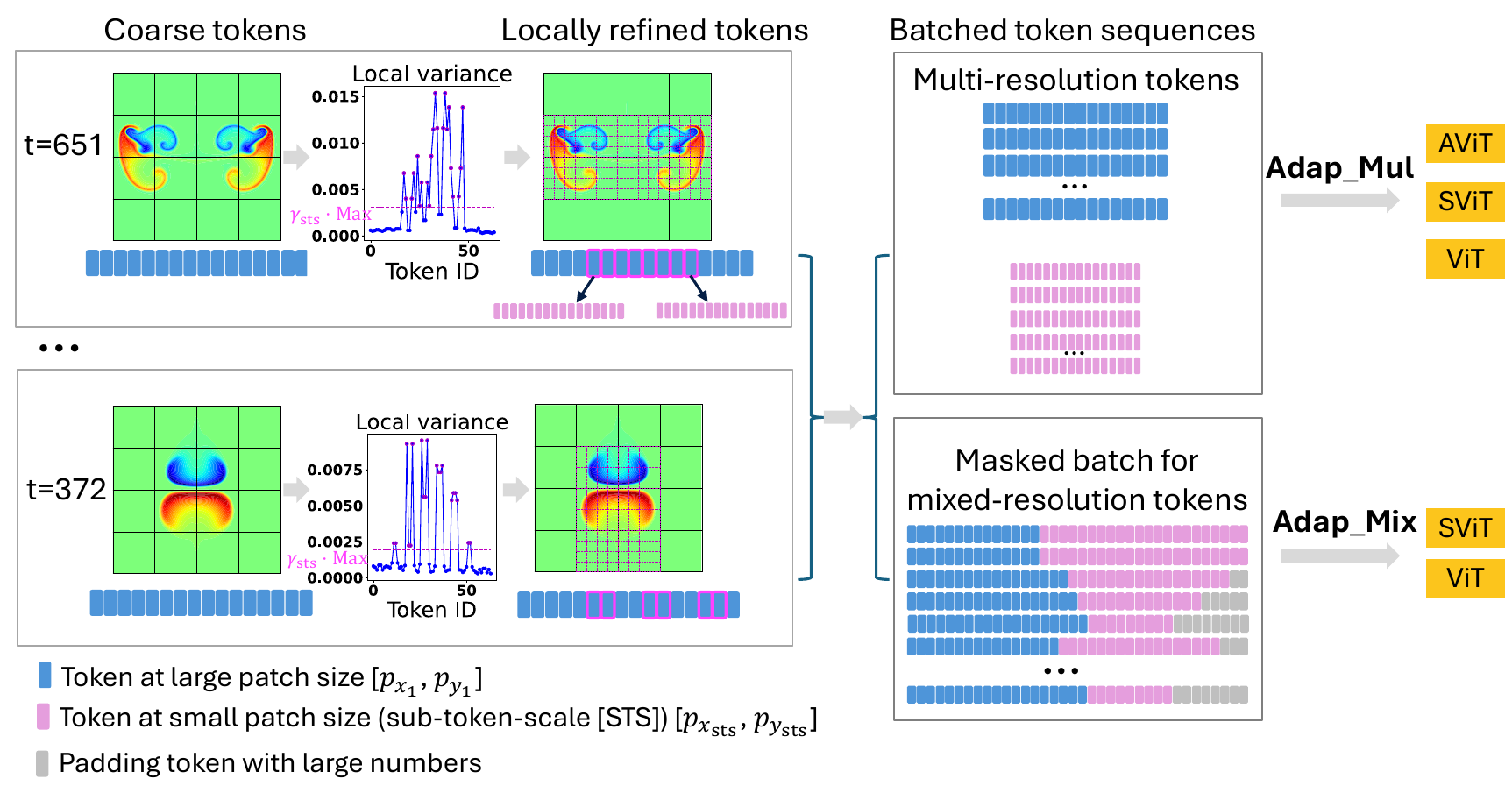}
\caption{Adaptive tokenization that dynamically adjusts patch sizes based on local features. There are three essential parameters: $[p_{x_1}, p_{y_1}]$, $[p_{x_{\textup{sts}}}, p_{y_{\textup{sts}}}]$ and $\gamma_{\textup{sts}}$. The parameter $[p_{x_1}, p_{y_1}]$ denotes the initial coarse patch size, $[p_{x_{\textup{sts}}}, p_{y_{\textup{sts}}}]$ represents the refined patch size, and $\gamma_\textup{sts}\in[0,1]$ determines which patches to refine. We select patches with local variances greater than $\gamma_\textup{sts}$ times the maximum variance across all patches (see Equation (\ref{eq-ind})).}
\label{fig-adaptoken}
\end{figure}

For a given solution field $\boldsymbol{u}_t\in \mathbb{R}^{H\times W \times C}$, tokenization at a constant patch size $[p_x, p_y]$ is achieved through a CNN block and leads to a patch grid of size $(npx, npy)=(H/p_x, W/p_y)$. 
For adaptive tokenization, we apply varying patch sizes in space based on local complexity represented by the patch variance.
For a solution $\boldsymbol{u}_t\in \mathbb{R}^{H\times W \times C}$ and an initial coarse patch size $[p_{x_1}, p_{y_1}]$, a variance tensor $\boldsymbol{v}_t \in \mathbb{R}^{npx_1\times npy_1}$ ($npx_1=H/p_{x_1}$ and $npy_1= W/p_{y_1}$) is calculated from solutions inside each patch of the reshaped solution $\widetilde{\boldsymbol{u}}_t \in \mathbb{R}^{npx_1\times npy_1\times p_{x_1} \times p_{y_1} \times C}$ as 
\begin{equation}\label{eq-variance}
    \boldsymbol{v}_t(i,j)=\frac{1}{C\cdot p_{x_1}\cdot p_{y_1}}\sum_{c=1}^C\sum_{k=1}^{p_{x_1}}\sum_{l=1}^{p_{y_1}}\left(\widetilde{\boldsymbol{u}}_t(i,j,k,l,c)-\frac{1}{p_{x_1}\cdot p_{y_1}}\sum_{k=1}^{p_{x_1}}\sum_{l=1}^{p_{y_1}}\widetilde{\boldsymbol{u}}_t(i,j,k,l,c)\right)^2.
\end{equation}
Patches with variance values greater than a prescribed threshold are then selected for further refinement at a smaller patch size. Specifically, let $\textup{STS-IDs}$ denote the index set of patches to be refined, then
\begin{equation}\label{eq-ind}
    \textup{STS-IDs}:=\{(i,j)|\boldsymbol{v}_t(i,j)>\gamma_{\textup{sts}}\cdot\boldsymbol{v}_{t,\max}\}, \quad N_\textup{sts}:=|\textup{STS-IDs}|,
\end{equation}
where $\gamma_{\textup{sts}}\in[0,1]$ is a user-specified hyperparameter, $\boldsymbol{v}_{t,\max}$ is the maximal variance among all patches, and $N_\textup{sts}$ is the number of patches to be refined.
The selected patches are refined to patches of a smaller size $[p_{x_{\textup{sts}}}, p_{y_{\textup{sts}}}]$, referred to as ``STS tokens'' in this work, where  ${\boldsymbol{Z}}^{ 0}_{\textup{sts},i}=\left[\boldsymbol{z}^{0}_{\textup{sts},1},\boldsymbol{z}^{0}_{\textup{sts},2},\ldots, \boldsymbol{z}^{0}_{\textup{sts},p_{x_1}/p_{x_\textup{sts}}\times p_{y_1}/p_{y_\textup{sts}}}\right]_i$ ($i=1,\dots,N_{\textup{sts}}$).
The STS tokens can be combined with the coarse tokens in two ways, as shown in Figure \ref{fig-adaptoken}. In the first approach, referred to as ``Adap\_Mul" (for adaptive multi-resolution tokenization), we consider the coarse and STS tokens as separate sequences, passing through the attention blocks serially.
In the second approach, referred to as ``Adap\_Mix" (for adaptive mixed-resolution tokenization), we replace the selected coarse patches with the sequence of STS tokens directly appended to the end of the sequence.

After spatiotemporal attention, the decoding of adaptive patch sequences into solution fields within the multi-physics postprocessor is performed using transposed convolutional blocks, tailored to each corresponding scale. 
For Adap\_Mul, the patches at different resolutions/sizes are deconvoluted separately and then summed to the final output, $\widehat{\boldsymbol{u}}_t$. Specifically, for a coarse attention output ${\boldsymbol{Z}}^L_\textup{coarse}=\left[\boldsymbol{z}^{L}_1,\boldsymbol{z}^L_2,\ldots, \boldsymbol{z}^L_{npx_1\times npy_1}\right]$ and STS attention outputs ${\boldsymbol{Z}}^{ L}_{\textup{sts},i}=\left[\boldsymbol{z}^{L}_{\textup{sts},1},\boldsymbol{z}^{L}_{\textup{sts}, 2},\ldots, \boldsymbol{z}^{L}_{p_{x_1}/p_{x_\textup{sts}}\times p_{y_1}/p_{y_\textup{sts}}}\right]_i$ ($i=1,\dots,N_{\textup{sts}}$), ``Adap\_Mul" performs the following operations:
\begin{equation}\label{eq-debed-adap-mul}
\begin{aligned} 
    \textup{Reconstruction from coarse patches:} \quad &\widehat{\boldsymbol{u}}_{t}=\textup{ConvTranspose2d}_1({\boldsymbol{Z}}^L_\textup{coarse}),
    \\
    \textup{Reconstruction from STS patches:} \quad     &\widehat{\boldsymbol{u}}_{t,\textup{sts},i}=\textup{ConvTranspose2d}_2({\boldsymbol{Z}}^{L}_{\textup{sts},i})\\
     &\widehat{\boldsymbol{u}}_{t,\textup{sts}}=[ \widehat{\boldsymbol{u}}_{t,\textup{sts},1},\ldots,  \widehat{\boldsymbol{u}}_{t,\textup{sts},N_\textup{sts}}] \\
    \textup{Fusion of multi-resolution solutions:} \quad
    &\widehat{\boldsymbol{u}}_t[\textup{STS-IDs}] = \widehat{\boldsymbol{u}}_{t}[\textup{STS-IDs}] + \widehat{\boldsymbol{u}}_{t,\textup{sts}}.
\end{aligned} 
\end{equation}

On the other hand, the Adap\_Mix approach fuses the coarse and STS patch sequences into the full sequences at the coarse and fine STS scales, respectively, reconstructs the solutions via transposed convolutions at corresponding resolutions separately, and then merges them to achieve multi-resolution solutions. This approach guarantees consistency with the coarse patch solution when $\gamma_\textup{sts}=1.0$ (no refinement) and the fine patch solution when $\gamma_\textup{sts}=0.0$ (refining all patches). 
Let ${\boldsymbol{Z}}^{\prime L}_\textup{coarse}=[\boldsymbol{z}^{L}_1,\boldsymbol{z}^L_2,\ldots, \boldsymbol{z}^L_{npx_1\times npy_1- N_\textup{sts}}]\in \mathbb{R}^{(npx_1\times npy_1- N_\textup{sts})\times C_\textup{emb}}$ denote the coarse portion of the mixed-resolution attention output, and let ${\boldsymbol{Z}}^{L}_{\textup{sts}, i}=[\boldsymbol{z}^{L}_1,\boldsymbol{z}^{L}_2,\ldots, \boldsymbol{z}^{L}_{p_{x_1}/p_{x_\textup{sts}}\times p_{y_1}/p_{y_\textup{sts}}}]_i$ ($i=1,\ldots, N_\textup{sts}$) denote the STS portion. Adap\_Mix performs the following operations:
\begin{enumerate}
    \item Reconstruct the full coarse patches $\boldsymbol{Z}^{L}_\textup{coarse}\in \mathbb{R}^{npx_1\times npy_1\times C_\textup{emb}}$ via 
    \begin{equation}
    \begin{aligned}
    {\boldsymbol{Z}}^{L}_\textup{coarse}[\textup{Kep-IDs}]&={\boldsymbol{Z}}^{\prime L}_{\textup{coarse}}, \\
    {\boldsymbol{Z}}^{L}_\textup{coarse}[\textup{STS-IDs}]&=[\textup{Mean}(\boldsymbol{Z}^{L}_{\textup{sts},1}),\textup{Mean}({\boldsymbol{Z}}^{L}_{\textup{sts}, 2}), \ldots,  \textup{Mean}({\boldsymbol{Z}}^{L}_{\textup{sts}, N_\textup{sts}})],
    \end{aligned}
    \end{equation}
    where $\textup{Kep-IDs}$ is the complementary indexing tensor to $\textup{STS-IDs}$, representing all coarse patches kept in the sequence.
    \item Reconstruct the full fine patches $\boldsymbol{Z}^{L}_\textup{fine}\in \mathbb{R}^{H/p_{x_\textup{sts}}\times W/p_{y_\textup{sts}}\times C_\textup{emb}}$ via %
    \begin{equation}
    \begin{aligned}
    {\boldsymbol{Z}}^{\prime L}_\textup{fine}[\textup{STS-IDs}, :,:]&=[{\boldsymbol{Z}}^{L}_{\textup{sts},1},{\boldsymbol{Z}}^{L}_{\textup{sts}, 2}, \ldots,  {\boldsymbol{Z}}^{L}_{\textup{sts}, N_\textup{sts}}],\\
    {\boldsymbol{Z}}^{\prime L}_\textup{fine}[\textup{Kep-IDs},:,:]&=\textup{repeat}\left({\boldsymbol{Z}}^{\prime L}_{\textup{coarse}}, {p_{x_1}/p_{x_\textup{sts}}\times p_{y_1}/p_{y_\textup{sts}}}\right), \\
     {\boldsymbol{Z}}^{ L}_\textup{fine}&=\textup{reshape}\left({\boldsymbol{Z}}^{\prime L}_\textup{fine}\right).
    \end{aligned}
    \end{equation}
    where ${\boldsymbol{Z}}^{\prime L}_\textup{fine} \in \mathbb{R}^{(npx_1\times npy_1)\times ({p_{x_1}/p_{x_\textup{sts}}\times p_{y_1}/p_{y_\textup{sts}}})\times C_\textup{emb}}$ is an intermediate supporting tensor.
   \item Reconstruct solution fields $\widehat{\boldsymbol{u}}_{t,\textup{coarse}}\in  \mathbb{R}^{H\times W \times C}$ and $\widehat{\boldsymbol{u}}_{t,\textup{fine}} \in  \mathbb{R}^{H\times W \times C}$ from coarse patches and fine patches, respectively: 
   \begin{equation}
   \begin{aligned}
   \widehat{\boldsymbol{u}}_{t,\textup{coarse}}=\textup{ConvTranspose2d}_1({\boldsymbol{Z}}^L_\textup{coarse}),\quad
   \widehat{\boldsymbol{u}}_{t,\textup{fine}}=\textup{ConvTranspose2d}_2({\boldsymbol{Z}}^{L}_{\textup{fine}}).
   \end{aligned}
   \end{equation}
    \item Fusion of solutions from step 3 to get the multi-resolution solution fields $\widehat{\boldsymbol{u}}_{t}\in  \mathbb{R}^{H\times W \times C}$ :
     \begin{equation}
    \begin{aligned}
   \widehat{\boldsymbol{u}}_t[\textup{Kep-IDs}] = \widehat{\boldsymbol{u}}_{t,\textup{coarse}}[\textup{Kep-IDs}], \quad
   \widehat{\boldsymbol{u}}_t[\textup{STS-IDs}] = \widehat{\boldsymbol{u}}_{t,\textup{fine}}[\textup{STS-IDs}].
    \end{aligned}
    \end{equation}
\end{enumerate}

Among the two adaptive approaches, Adap\_Mul is simpler to implement, requiring minimal code modifications, supports the AViT attention mechanism, and does not increase the maximum sequence lengths. In contrast, Adap\_Mix produces relatively longer sequences and lacks AViT support but has the potential significant benefit of better capturing cross-scale correlations than the decoupled Adap\_Mul.
Furthermore, by varying $\gamma_{\textup{sts}}$ from 1.0 to 0.0, Adap\_Mix guarantees a smooth transition from the coarse patch solution at $[p_{x_1}, p_{y_1}]$ to the fine patch solution at $[p_{x_{\textup{sts}}}, p_{y_{\textup{sts}}}]$ (see Figure \ref{fig-adapdummy-gamma-adapmix}).

\paragraph{Pretraining and fine-tuning.} We pretrain the models on PDEBench data, which includes five basic 2D systems: incompressible flows, compressible flows, turbulent flows, reaction-diffusion systems, and shallow water equations. We consider two fine-tuning cases: 1) colliding thermals between a cold and a warm bubbles from MiniWeather simulations \citep{norman2020miniweather} and 2) lid-driven cavity MHD flows \citep{FAMBRI2023112493}. As discussed in detail in Appendix~\ref{app-data}, these fine-tuning datasets were selected to be meaningfully out-of-distribution, not only in flow regime but also in including thermal and electromagnetic components that are not represented at all in the pretraining data. Training was performed on the Frontier and Perlmutter supercomputers at the Oak Ridge Leadership Computing Facility (OLCF) and National Energy Research Scientific Computing Center (NERSC), respectively.

\section{Experiments}\label{sec-exp}

We design three experiments to evaluate 1) the performance of three spatiotemporal attention schemes, AViT, SViT, and ViT, 2) the impact of adaptive tokenization, and 3) the effectiveness of pretrained models on two fine-tuning tasks that feature physics different from the pretraining data.
In these experiments, we set $p_t=1$ and $C_\textup{uni}=C_\textup{emb}/4$, and employ square patches (i.e., $p_x=p_y$, $p_{x_1}=p_{y_1}$, and $p_{x_{\textup{sts}}}=p_{y_{\textup{sts}}}$) by default. 

\subsection{Spatiotemporal attention schemes} \label{sec-atts}

We evaluate AViT, SViT, and ViT for three model sizes: Tiny (Ti), Small (S), and Base (B) with 3, 6, 12 heads and hidden dimension $C_{\textup{emb}} =$ 192, 384, and 768, respectively \citep{touvron2022three}, as shown in Table \ref{tab-modelsize}, on the colliding thermals dataset.
 In the same size category, AViT and SViT are about 30\% larger than ViT due to the additional attention block.
More details about the experiment are presented in Appendix~\ref{app-sec-algorithm}.

\begin{table}[h!]
\caption{Number of model parameters in AViT, SViT, and ViT for three model sizes, Tiny, Small, and Base, detailed in Section~\ref{sec-atts}. ViT results in about 30\% fewer model parameters than AViT and SViT because the latter two require additional attention blocks.}\label{tab-modelsize}
  \begin{center}
    \begin{tabular}{r|c|c|c} 
      \textbf{ }&Tiny & Small & Base  \\
      \hline
      AViT&7.5M&29.9M&119.3M\\
      SViT&7.6M&30.0M&119.3M\\
      ViT&5.8M&22.8M&90.9M\\
    \end{tabular}
  \end{center}
\end{table}

\begin{figure}[h]
\centering
\includegraphics[width=0.95\textwidth]{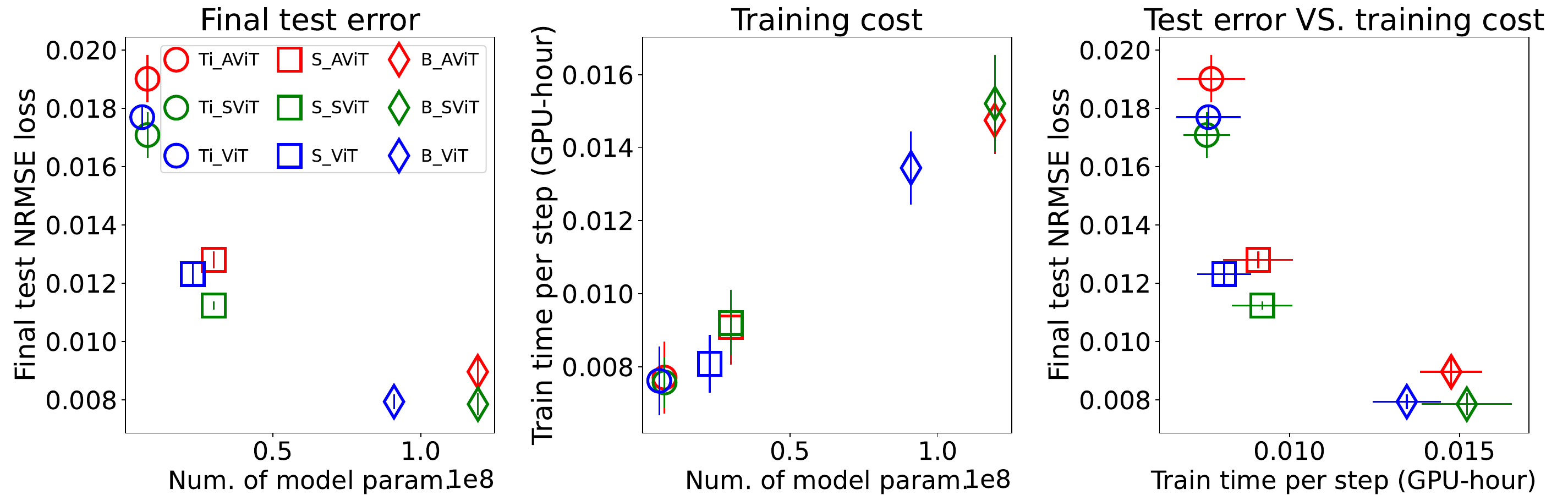}
\caption{Learning efficiency of AViT, SViT, and ViT at three model sizes regarding final predictive error and training time cost: SViT and ViT are observed to be more expressive and computationally efficient than AViT in the experiment, as they require fewer model parameters and less training time to achieve the same test accuracy. 
}
\label{fig-att}
\end{figure}

Figure~\ref{fig-att} compares the final test error, defined as the normalized root-mean-square error (NRMSE), and the training time, represented as GPU hour per step, for the nine models.
For the same size category, SViT (green) achieves the lowest error, followed by ViT (blue), and then AViT (red).
In terms of training time, SViT takes longer than AViT, while ViT is the least expensive.
ViT processes longer token sequences and hence is expected to have a higher single-unit attention cost, whereas AViT and SViT have multiple attention units with shorter token sequence lengths. 
The results reported in Figure~\ref{fig-att} show that the ViT has the lowest cost, which implies that the number of attention blocks plays a more important role than the token sequence length in terms of training cost in this example. 
This observation is due to the fact that the spatiotemporal token sequence length ($16\times 8 \times 8$) in this example is relatively short. We expect ViT to become more expensive than AViT and SViT when more refined or higher dimensional solutions are considered, in which longer token sequences are required. 
In general, we find that SViTs and ViTs are more expressive and computationally efficient than AViTs, in that they achieve the lower predictive errors with fewer model parameters and less training time.

\subsection{Adaptive tokenization}\label{sec-adap}

\begin{figure}[h]
\centering
\includegraphics[width=0.95\textwidth]{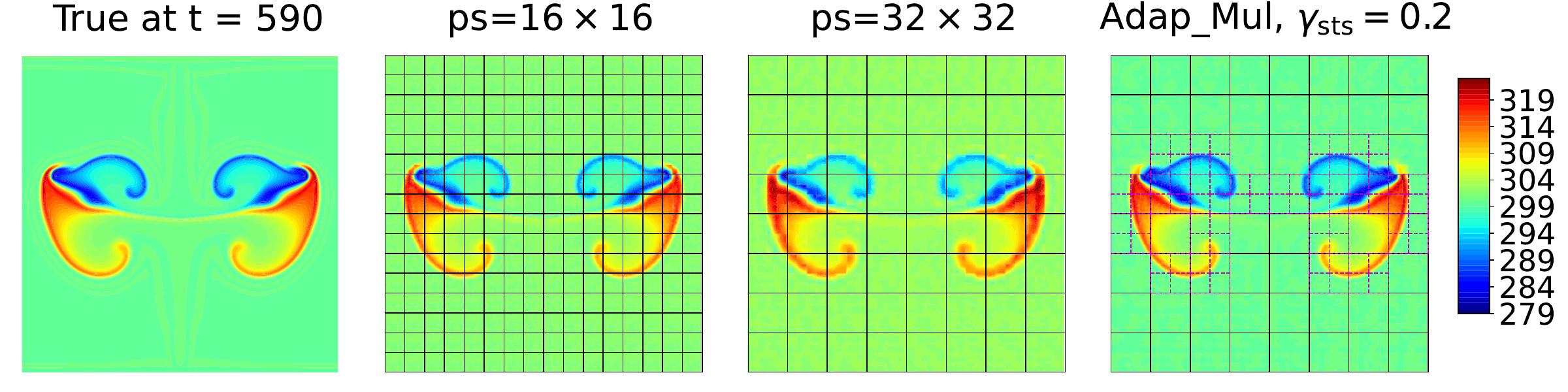}
\caption{Predicted temperature contours at $t=590$ from Ti-SViT models with constant patch sizes ps=$16\times16$ and ps=$32\times32$ and adaptive tokenization (Adap\_Mul with $p_{x_1} =p_{y_1}=32$, $p_{x_{\textup{sts}}}=p_{y_{\textup{sts}}}=16$ , and $\gamma_\textup{sts}=0.2$). Adap\_Mul predicts smoother, finer local structures that are overlooked in ps=$32\times32$, similar to the more expensive ps=$16\times16$.}
\label{fig-adapdummy-visual}
\end{figure}

We start the evaluation of our adaptive tokenization methods in a single collision trajectory between two thermal bubbles. Figure~\ref{fig-adapdummy-visual} compares the temperature contours of the true solution at $t=590$ with the predicted solutions from Ti-SViT models at constant patch sizes, ps=$16\times16$ and ps=$32\times32$, and adaptive tokenization (Adap\_Mul with $p_{x_1} =p_{y_1}=32$, $p_{x_{\textup{sts}}}=p_{y_{\textup{sts}}}=16$ , and $\gamma_\textup{sts}=0.2$).
The predicted solution from ps=$32\times32$ exhibits abrupt changes with clear edges for the local structures inside the patches, while the finer resolution model at ps=$16\times16$ captures smoother, finer structures but requires many more patches. In contrast, our adaptive tokenization methods capture smooth, fine structures comparable to ps=$16\times16$ while requiring much shorter sequences.

\begin{figure}[h]
\centering
\includegraphics[width=0.48\textwidth]{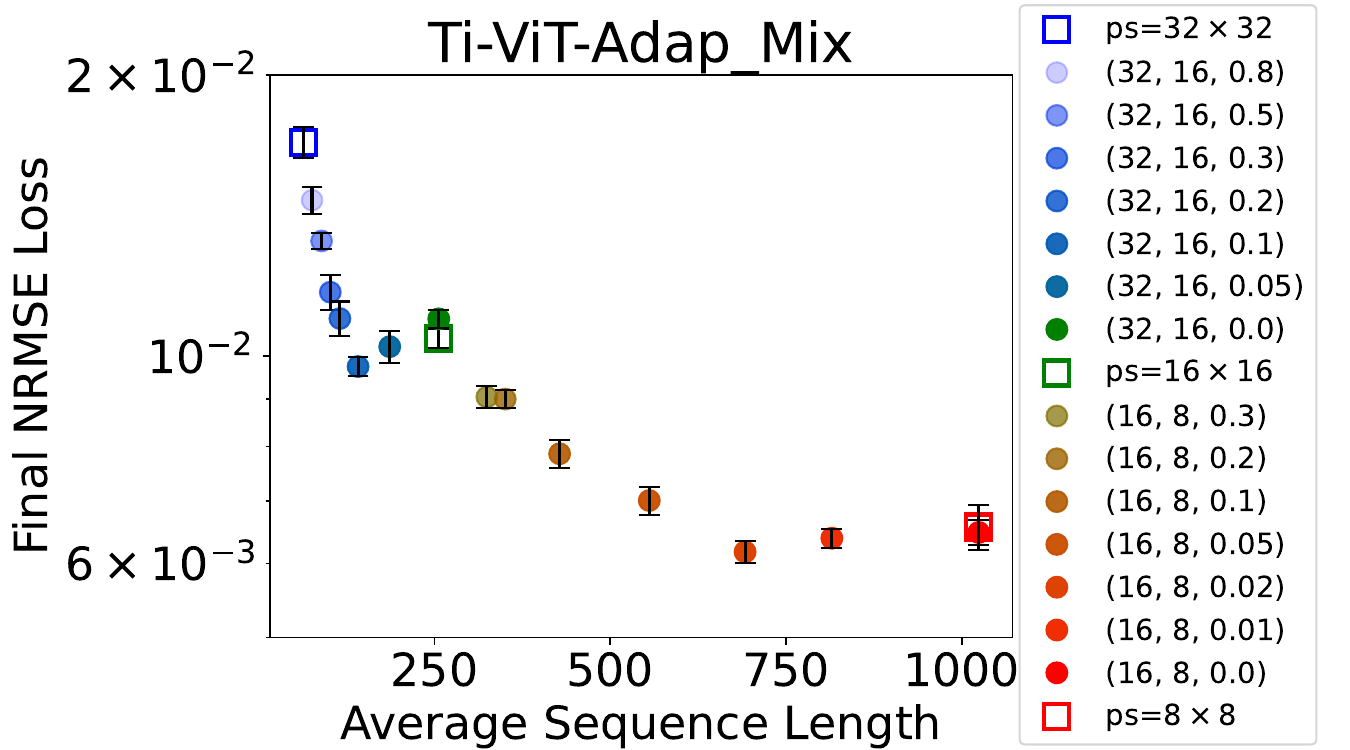}
\includegraphics[width=0.48\textwidth]{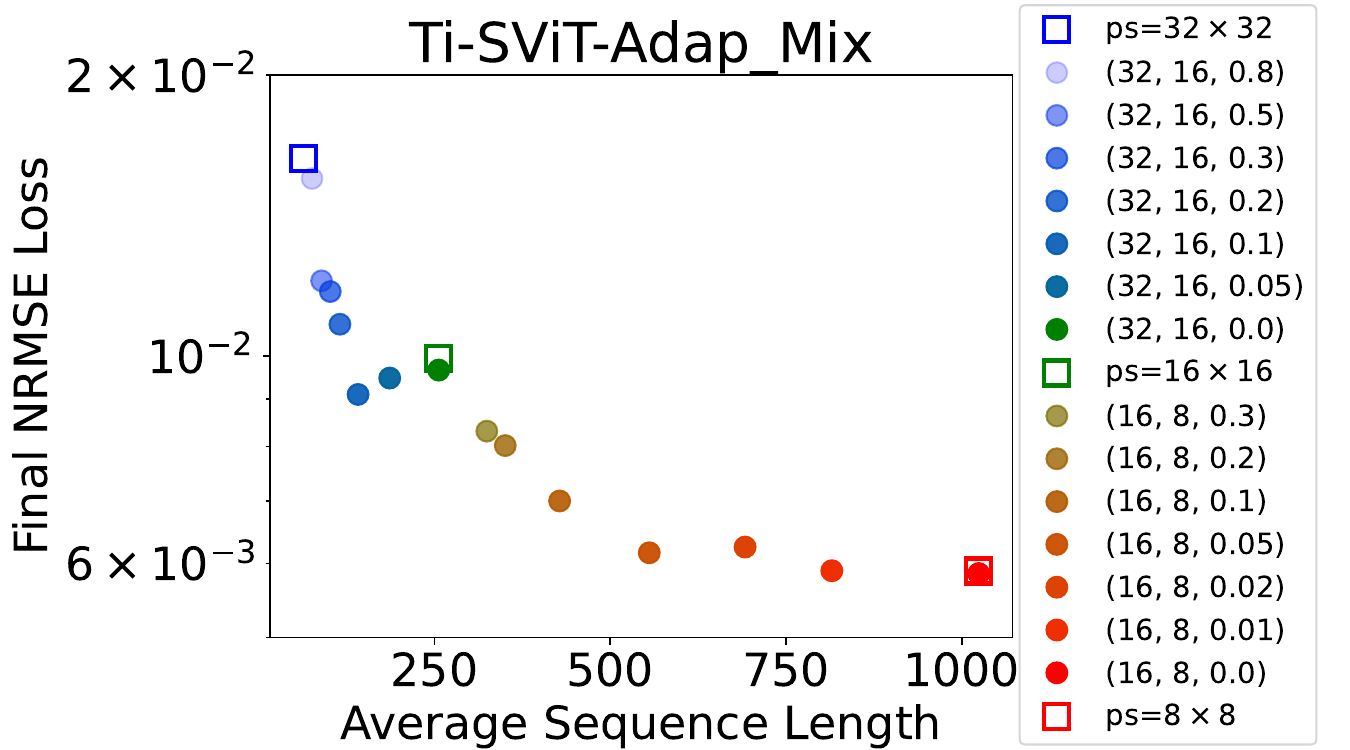}
\caption{Final NRMSE loss for Tiny ViT (left) and SViT (right) with adaptive tokenization --- \textbf{Adap\_Mix} with hyperparamters ($p_{x_1}$, $p_{x_{\textup{sts}}}$, $\gamma_\textup{sts}$)--- and constant patch sizes against average sequence length, $L_\textup{avg,mix}$ (Equation (\ref{eq-lmix})). Error bars, representing standard deviations from 3 runs, are shown for ViT.
Adap\_Mix with  $\gamma_\textup{sts}$ varying from 1.0 to 0.0 shows a clear convergent transition from the coarse constant patch size ps=$p_{x_1}\times p_{y_1}$ to the fine constant patch size ps=$p_{x_{\textup{sts}}}\times p_{y_{\textup{sts}}}$. More interestingly, Adap\_Mix is shown to achieve lower prediction errors than the more expensive ps=$p_{x_{\textup{sts}}}\times p_{x_{\textup{sts}}}$ cases despite requiring only half of the average sequence length.}
\label{fig-adapdummy-gamma-adapmix}
\end{figure}

\paragraph{Adap\_Mix in ViT and SViT } Adap\_Mix with ($p_{x_1}$, $p_{x_{\textup{sts}}}$, $\gamma_\textup{sts}$) is designed to ensure convergence in $\gamma_\textup{sts}$ values. When $\gamma_\textup{sts}\to 1$, no refinement is conducted and the output is converged to ps=$p_{x_1}\times p_{x_1}$. Conversely, when $\gamma_\textup{sts}\to 0$, all patches are refined and the output is converged to ps=$p_{x_{\textup{sts}}}\times p_{x_{\textup{sts}}}$.
To examine such convergence behavior, we conduct a set of runs with varying ($p_{x_1}$, $p_{x_{\textup{sts}}}$, $\gamma_\textup{sts}$) values, together with runs at constant patch sizes, ps=$32\times32$, ps=$16\times16$, and ps=$8\times8$.
Figure \ref{fig-adapdummy-gamma-adapmix} shows the final NRMSE test loss versus the average sequence length of patches per time step, $L_\textup{avg,mix}$. For a given trajectory of spatiotemporal solutions with $T$ steps, the average sequence length is defined as
\begin{equation}\label{eq-lmix}
    L_\textup{avg,mix} =\frac{1}{T} \sum_{t=1}^T  L_t=\frac{1}{T} \sum_{t=1}^T\left[\left(npx_1\cdot npy_1- N_{\textup{sts},t}\right) + N_{\textup{sts},t} \cdot \left({\frac{p_{x_1}}{p_{x_\textup{sts}}}\cdot \frac{p_{y_1}}{p_{y_\textup{sts}}}} \right)\right],
\end{equation}
where $L_t$ represents the length of the mixed patch sequence and $N_{\textup{sts},t}$ is the number of patches selected based on Equation (\ref{eq-ind}) at time $t$.
Clearly, the predictive error of Adap\_Mix evolves from the corresponding coarse patch results of ps=$p_{x_1}\times p_{x_1}$ to ps=$p_{x_{\textup{sts}}}\times p_{x_{\textup{sts}}}$ when $\gamma_\textup{sts}$ varying from 1.0 to 0.0, in two settings ($p_{x_1}=32, p_{x_{\textup{sts}}}=16$) and ($p_{x_1}=16, p_{x_{\textup{sts}}}=8$) and for both ViT and SViT.
More interestingly, Adap\_Mix at some $\gamma_\textup{sts}$ values even achieves a lower predictive error than the fine patch case ps=$p_{x_{\textup{sts}}}\times p_{x_{\textup{sts}}}$ with a much shorter  sequence length (e.g., with $2\times$ reduction), clearly indicating the advantages of the adaptive tokenization approach.

\begin{figure}[h]
\centering
\includegraphics[width=0.65\textwidth]
{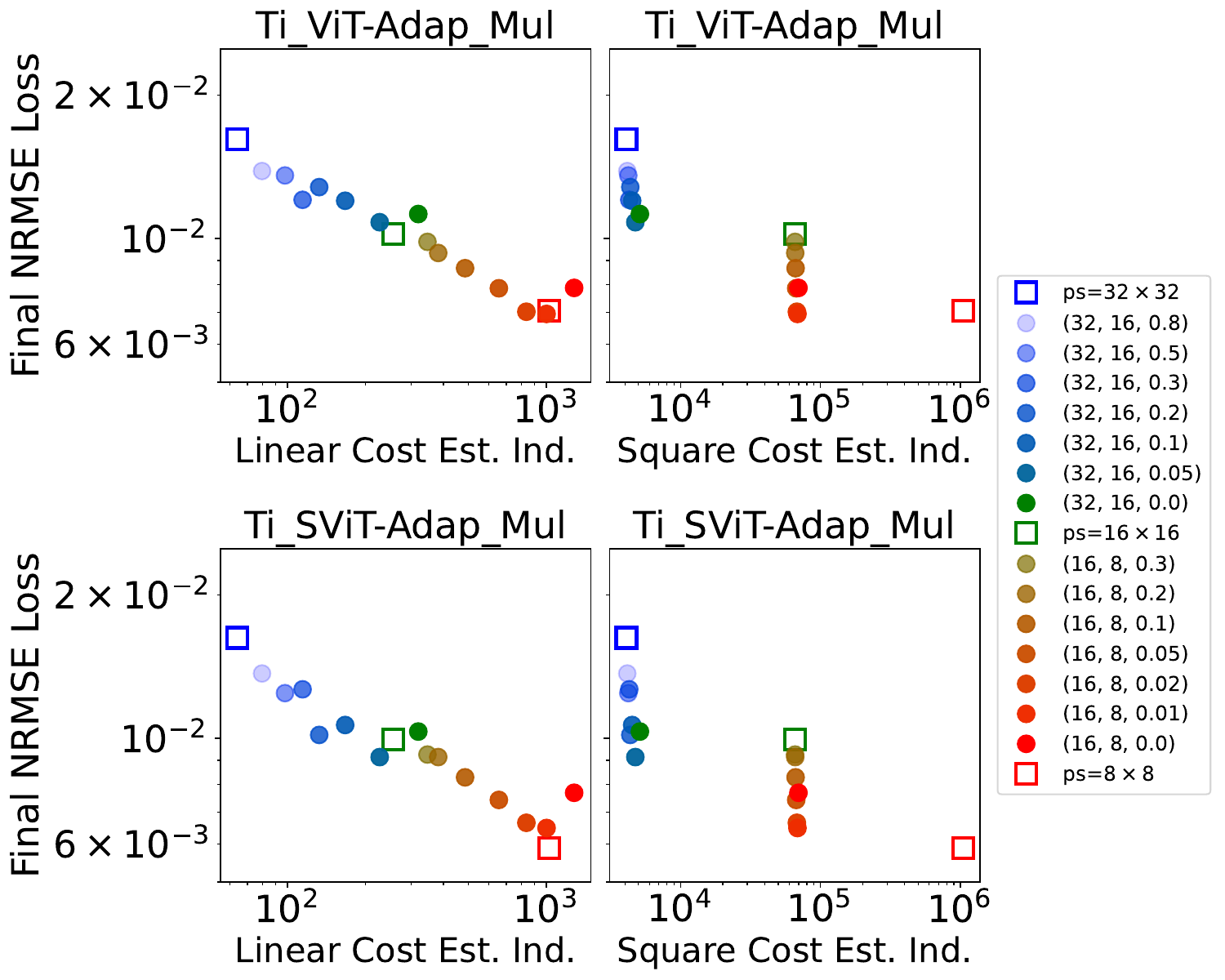}
\caption{Final NRMSE loss for Tiny ViT (top) and SViT (bottom)
with adaptive tokenization --- \textbf{Adap\_Mul} with hyperparamters ($p_{x_1}$, $p_{x_{\textup{sts}}}$, $\gamma_\textup{sts}$) --- and constant patch sizes against linear cost estimation index (left) and square cost estimation index (right). 
As $\gamma_\textup{sts}$ moving from 1.0 to 0.0, predictive errors decrease dramatically; simultaneously, the linear cost in \textbf{Adap\_Mul} transitions from ps=$p_{x_1}\times p_{y_1}$ to ps=$p_{x_{\textup{sts}}}\times p_{y_{\textup{sts}}}$, while quadratic cost increase is negligible.}
\label{fig-adapdummy-vit-gamma-conv}
\end{figure}

\begin{figure}[h]
\centering
\includegraphics[width=0.65\textwidth]
{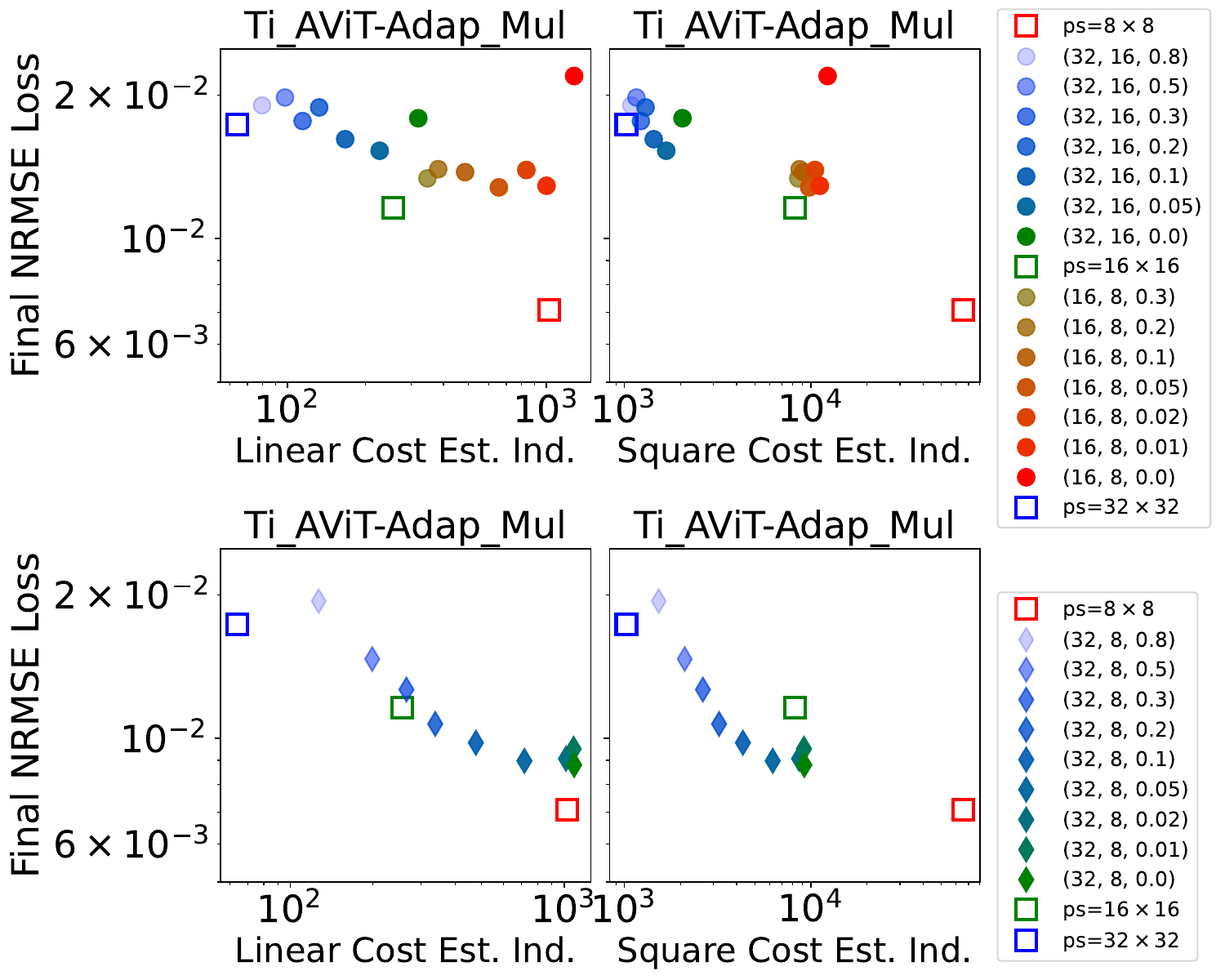}
\caption{Final NRMSE loss for Tiny AViT with adaptive tokenization --- \textbf{Adap\_Mul} with hyperparamters ($p_{x_1}$, $p_{x_{\textup{sts}}}$, $\gamma_\textup{sts}$)m--- and constant patch sizes against linear cost estimation index (left) and square cost estimation index (right). 
Compared with ViT and SViT, AViT requires smaller STS patches to achieve similar accuracy improvements due to axial decomposition, demonstrating notable prediction accuracy gains with $p_{x_{\textup{sts}}}=8$ instead of $p_{x_{\textup{sts}}}=16$, for  $p_{x_1}=32$.}
\label{fig-adapdummy-vit-gamma-conv-avit}
\end{figure}

\paragraph{Adap\_Mul in ViT, SViT, and AViT} In contrast to Adap\_Mix, which combines the coarse and locally refined patches into a hybrid sequence that is fed into attention blocks together, Adap\_Mul treats the two, i.e., ${\boldsymbol{Z}}^L_\textup{coarse}=\left[\boldsymbol{z}^{L}_1,\boldsymbol{z}^L_2,\ldots, \boldsymbol{z}^L_{npx_1\times npy_1}\right]$ and ${\boldsymbol{Z}}^{ L}_{\textup{sts},i}=\left[\boldsymbol{z}^{L}_{\textup{sts},1},\boldsymbol{z}^{L}_{\textup{sts}, 2},\ldots, \boldsymbol{z}^{L}_{p_{x_1}/p_{x_\textup{sts}}\times p_{y_1}/p_{y_\textup{sts}}}\right]_i$ ($i=1,\dots,N_{\textup{sts}}$), separately. Adap\_Mul maintains this separation for both attention and solution reconstruction and views the reconstructed solutions from the refined patches as a local STS correction. 
The computing cost scales either linearly for MLP or quadratically for attention with sequence length in various model components.
To represent the cost, we define the linear and quadratic indices for ViT and SViT as in
\begin{equation}\label{eq-lmul}
\begin{aligned}
       L_\textup{lin} &= \frac{1}{T}\sum_{t=1}^T  L_t= \frac{1}{T}\sum_{t=1}^T\left[npx_1\cdot npy_1 + N_{\textup{sts},t} \cdot \left({\frac{p_{x_1}}{p_{x_\textup{sts}}}\cdot \frac{p_{y_1}}{p_{y_\textup{sts}}}} \right)\right],\\
     L_\textup{quad} &= \frac{1}{T}\sum_{t=1}^T  L_{\textup{quad},t}= \frac{1}{T}\sum_{t=1}^T\left[\left(npx_1\cdot npy_1\right)^2 + N_{\textup{sts},t} \cdot \left({\frac{p_{x_1}}{p_{x_\textup{sts}}}\cdot \frac{p_{y_1}}{p_{y_\textup{sts}}}} \right)^2\right].
\end{aligned}
\end{equation}
For AViT, the index $L_{\textup{quad},t}$ needs to be adjusted as
\begin{equation}\label{eq-lmul-avit}
\begin{aligned}
     L_{\textup{quad},t} =  \left(npx_1^2\cdot npy_1+npx_1\cdot npy_1^2\right) + 
     N_{\textup{sts},t} \cdot \left[\left(\frac{p_{x_1}}{p_{x_\textup{sts}}}\right)^2\cdot \frac{p_{y_1}}{p_{y_\textup{sts}}}+\frac{p_{x_1}}{p_{x_\textup{sts}}}\cdot \left(\frac{p_{y_1}}{p_{y_\textup{sts}}} \right)^2\right],
\end{aligned}
\end{equation}
Figure \ref{fig-adapdummy-vit-gamma-conv} shows the final NRMSE test loss against the two cost estimation indices $L_\textup{lin}$ (left) and $L_\textup{quad}$ (right) of Adap\_Mul in ViT (top) and SViT (bottom) 
at varying values of ($p_{x_1}$, $p_{x_{\textup{sts}}}$, $\gamma_\textup{sts}$). Clearly, as $\gamma_\textup{sts}$ decreases for ViT and SViT, predictive errors are significantly reduced. Simultaneously, the linear cost increases progressively from coarse to refined and beyond, while the increase in quadratic cost remains negligible. Moreover, the accuracy does not converge to the refined case when $\gamma_\textup{sts}=0$, despite the higher linear cost. This indicates that Adap\_Mul is better suited for scenarios involving attention on long sequences.
Figure \ref{fig-adapdummy-vit-gamma-conv-avit} presents similar results for Adap\_Mul in AViT. Compared with ViT and SViT, the accuracy improvement in AViT is less pronounced in our experiments with $(p_{x_1}, p_{x_{\textup{sts}}}) = (32,16)\textup{ and }(16,8)$ (top), possibly due to the extremely short sequence lengths of 2 in AViT.
Notably, significant accuracy gains are observed when $p_{x_{\textup{sts}}}$ is reduced from 16 to 8 for $p_{x_1}=32$ (bottom). 

Comparing the two approaches for adaptive tokenization, we find that \textbf{Adap\_Mix} provides better predictive accuracy, likely due to considering cross-scale correlations, and guarantees convergence toward the solution with uniformly refined tokens. In contrast, \textbf{Adap\_Mul} is dramatically more cost effective for attention operations with quadratic complexity and easier to implement than Adap\_Mix. AViT does not interact well with adaptive tokenization approaches when the STS sequence length $p_{x_1}/p_{x_{\textup{sts}}}$ is too short.

\subsection{Effectiveness of pretraining in colliding thermals and MHD fine-tuning tasks} \label{sec-effect-pre}

We examine the transferrability of pretrained models to fine-tuning systems with distinct physics and different set of variables, as in Table \ref{tab-app-data}. Specifically, we aim to address three broad questions:
\begin{enumerate}
    \item Is pretraining effective when the downstream tasks have a distinct set of physical variables?
    \item How does limited fine-tuning of non-attention blocks compare to full fine-tuning?
    \item  How does the amount of fine-tuning data affect convergence?
\end{enumerate}
To address these three questions, we design a sets of experiments, starting from models pretrained on PDEBench or randomly initialized models (``*\_INIT''), and fine-tune them on colliding thermals and MHD datasets with distinct physical variables. For fine-tuning each model, we either allow all model parameters to be tunable (``ALL'') or freeze the attention blocks and limit training to the preprocessor, the tokenization module, and the postprocessor (``PREPOST''). Finally, for each initial model and fine-tuning configuration, we train four models with increasing amounts of fine-tuning data.

\begin{figure}[h]
\centering
\includegraphics[width=0.4\textwidth]{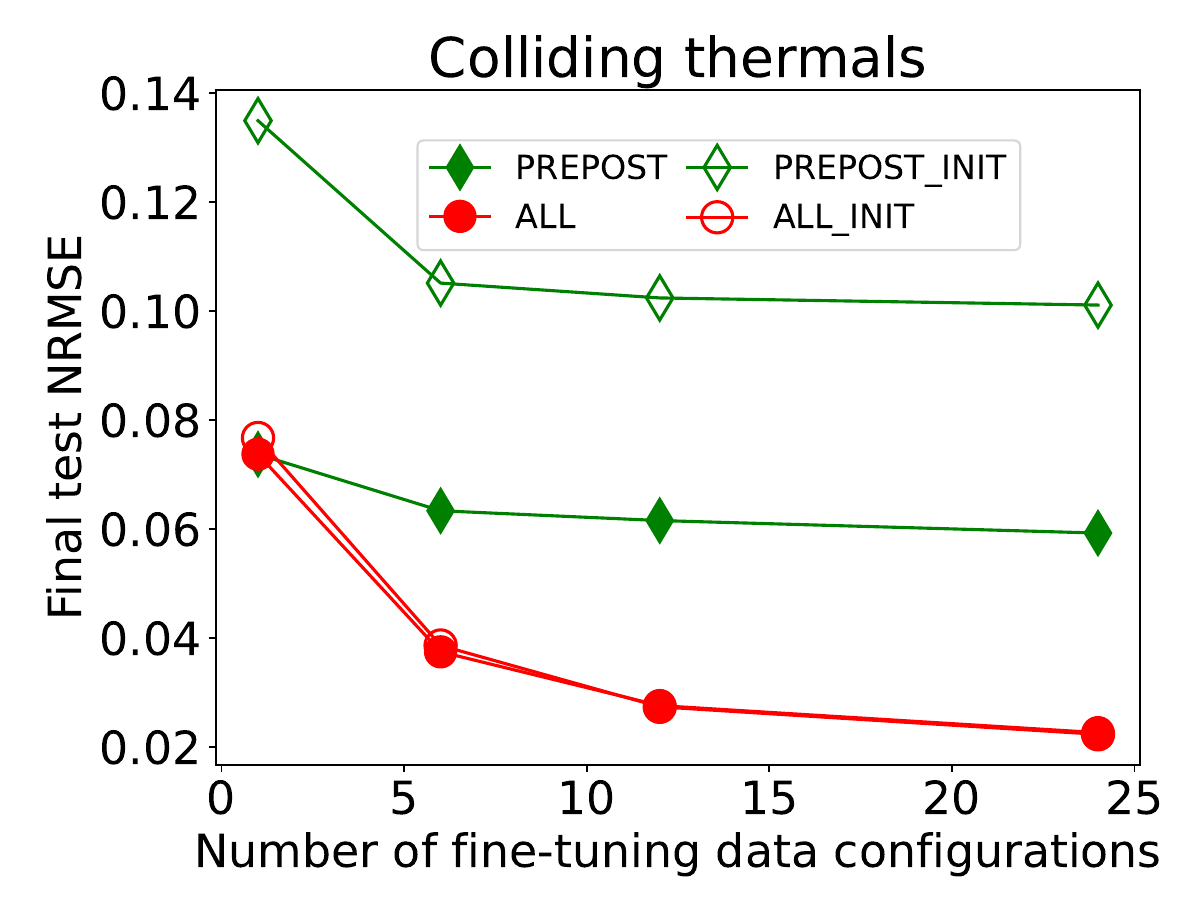}
\caption{NRMSE loss for test set at different training data sizes in fine-tuning of colliding thermals at a maximum lead time of 50 steps, with full (``ALL'') and limited (``PREPOST'') fine-tuning using pretrained and randomly initialized models (``*\_INIT'').}
\label{fig-MW}
\end{figure}

\begin{figure}[h]
\centering
\includegraphics[width=0.95\textwidth]{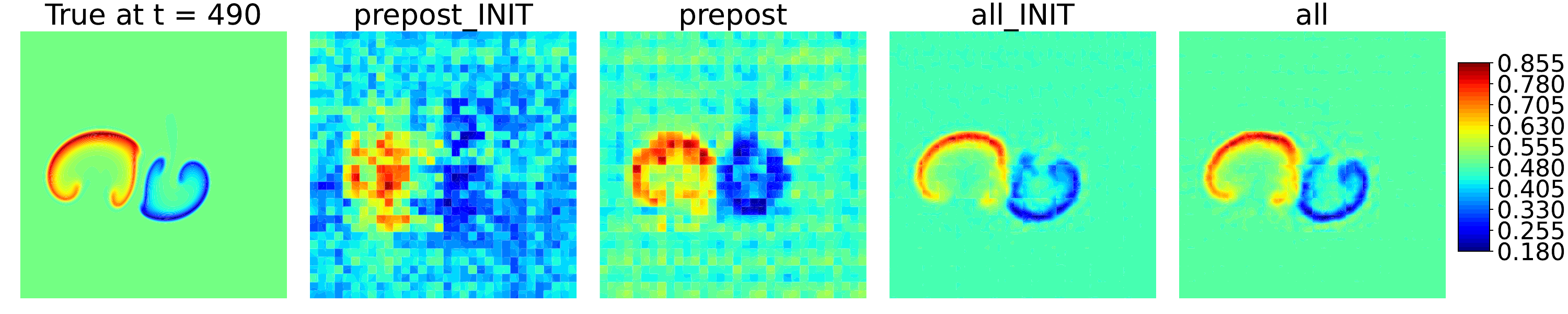}
\caption{Temperature contours of true solution vs predicted solutions from four fine-tuned models (on 12 trajectories) at $t=490$ from Ti-SViT models for a lead time of 40 in the collision of two thermal bubbles.}
\label{fig-finet-MW-visual}
\end{figure}

For the colliding thermals dataset, Figure~\ref{fig-MW} compares the test loss with full and limited fine-tuning using pretrained and randomly initialized models. The different training data sizes ranging from one set of colliding thermals time-trajectory to 24 sets of trajectories. The fine-tuning task is to predict the solution of the physical system at a lead time of $t_{\textup{lead}}$ uniformly sampled between 1 and 50 steps. An example of the true and predicted solutions in these four training configurations is illustrated in Figure~\ref{fig-finet-MW-visual}.

For the limited fine-tuning test with the colliding thermals dataset, the pretrained models achieve significantly lower error than starting from scratch with randomly initialized parameters. Moreover, while this advantage persists as the number of fine-tuning data increases,
it is most pronounced in the low data configuration of learning from a single trajectory. Indeed, we find that limited fine-tuning with the pretrained models generalizes well even when learning from one trajectory, seeing only moderate improvements when run on the largest dataset size considered. Overall, the lower converged error from pretrained models suggests the frozen attention blocks clearly learned transferable knowledge during pretraining. For full fine-tuning, the accuracy is much better than limited fine-tuning as a result of the model being more expressive. The difference between the pretrained and randomly initialized models is much lower, being minor in the case of a single data configuration during training and vanishing as the amount of data increases. 

\begin{figure}[h]
\centering
\includegraphics[width=0.35\textwidth]{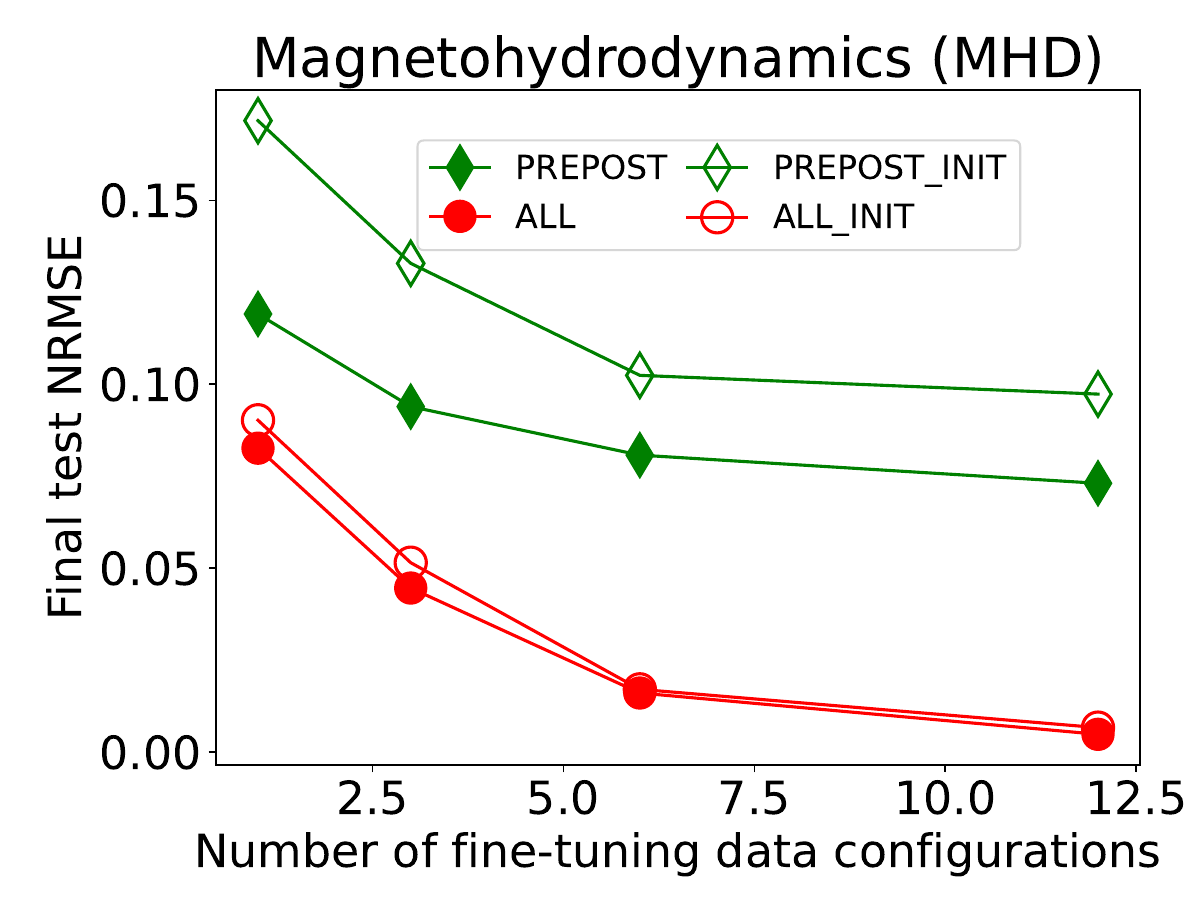}
\vspace{-0.25cm}
\caption{NRMSE loss for test set at different training data sizes in fine-tuning of lid-driven cavity MHD flows dataset at a maximum lead time of 100 steps, with full (``ALL'') and limited (``PREPOST'') fine-tuning using pretrained and randomly initialized models (``*\_INIT'').}
\label{fig-MHD}
\end{figure}

\begin{figure}[h]
\centering
\includegraphics[width=0.85\textwidth]{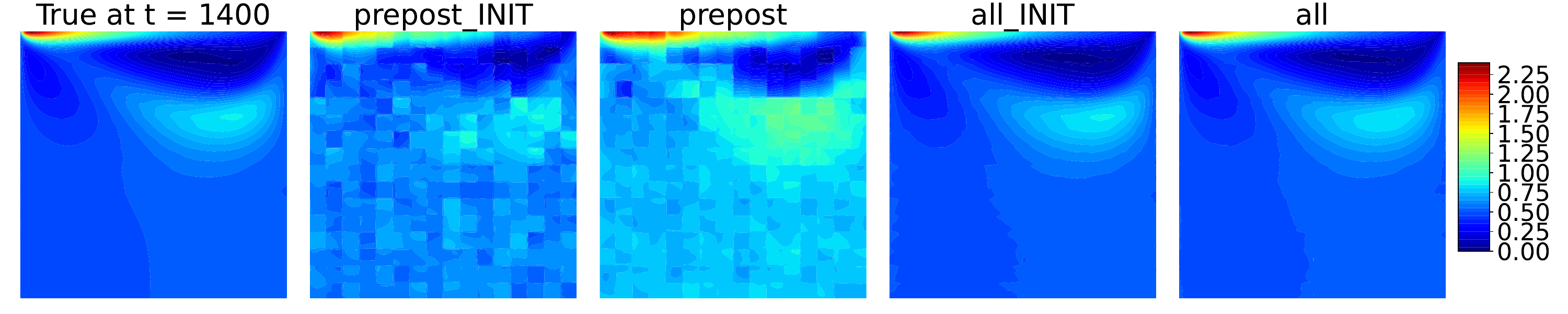}
\caption{Contours of true horizontal magnetic field values $B_x$ vs predicted solutions from four fine-tuned models (on 12 trajectories) at $t=1400$ from Ti-SViT models for a lead time of 80 in lid-driven cavity MHD flows.}
\label{fig-finet-MHD-visual}
\end{figure}

For the MHD dataset, Figure~\ref{fig-MHD} shows the final test NRMSE errors in lid-driven cavity flows after fine-tuning against data sizes when starting from pretrained and randomly initialized models for limited and full fine-tuning. The training dataset sizes used for fine-tuning range from 1 to 12 simulation configurations, with each configuration including approximately 1900 samples. The fine-tuning task is to predict the flow solution at a lead time of $t_{\textup{lead}}$ uniformly sampled between 1 and 100 steps. Contour plots from the true solution and the predicted solution from each training configuration are depicted in Figure \ref{fig-finet-MHD-visual}.

Overall, the fine-tuning performance is a result of model expressibility, training data size, and the similarity between training and testing tasks. As with the colliding thermals dataset, pretrained models outperformed the randomly initialized models for both full and limited fine-tuning runs. However, the reduced expressibility of the limited fine-tuning configuration consistently shows an accuracy gap, even with more training data, as they cannot fully represent the data complexity. In contrast, full fine-tuning leads to more expressive models that can capture all training data information when trained on limited data but often show high test errors; as more training data is provided, they generalize well and lead to a convergent improved test error. In our fine-tuning, the randomly initialized models perform well in testing even with a single data configuration, likely due to the similarity between training and testing tasks. Future work will explore more challenging scenarios with increased heterogeneity within the fine-tuning data.

While studies like \cite{mccabe2023multiple} have demonstrated impressive outperformance from fine-tuning of pretrained models versus randomly initialized models, these fine-tuning tests were performed on data that, while distinct, was fully governed by physical equations and characterized by physical variables that were represented in the training data. Yet for a model that aims to be foundational for multiphysical systems, we argue that assessing model performance in more realistic settings, where equations like Navier-Stokes are coupled with those from other domains of physics, is a more informative test of the effectiveness of pretraining. Accordingly, we assess fine-tuning performance on physical systems that incorporate fluid flows, which are well-represented in PDEBench, with thermodynamics and electromagnetism, which are not. As reasonably anticipated, we find that advantages of pretraining are reduced in this more complex setting.

\section{Discussion}\label{sec-discussion}

In this paper, we make three contributions that will advance the development of foundation models for multiscale physical systems. First, we find that while some data efficiency is lost in a fully decoupled spatiotemporal attention scheme such as AViT, SViT provides an intriguing balance of computational and data efficiency versus the standard ViT approach. Yet using SViT alone does not sufficiently address the computational challenges associated with attention for high spatial resolutions. Second, we instead suggest that our adaptive tokenization scheme provides a promising approach for working with high resolution data. This sort of adaptivity has the potential to be both flexible and expressive enough to deal with the dynamic and sparse nature of the multiscale features in physical data. Third, we suggest an alternative path to evaluate foundation models for multiscale physical systems that focuses on fine-tuning problems involving out-of-distribution physics governed by different equations with distinct sets of physical variables. In two such settings, colliding thermals and magnetohydrodynamics, we find that while pretraining does provide an advantage, its impact is much more muted compared to fine-tuning on the same set of variables, suggesting additional effort is required to obtain truly foundational models in this space.

\section*{Data and software availability}

We will publicly release the data, code, and trained models upon the publication of this paper.

\section*{Acknowledgments}
This research is sponsored by the Artificial Intelligence Initiative as part of the Laboratory Directed Research and Development (LDRD) Program of Oak Ridge National Laboratory, managed by UT-Battelle, LLC, for the US Department of Energy under contract DE-AC05-00OR22725. This material is based in part upon work carried out in the ‘Center for Simulation of Plasma - Liquid Metal Interactions in Plasma Facing Components and Breeding Blankets of a Fusion Power Reactor’ project, supported by the U.S. Department of Energy, Office of Science, Office of Advanced Scientific Computing Research and Office of Fusion Energy Sciences, Scientific Discovery through Advanced Computing (SciDAC) program. 

This research used resources of the Oak Ridge Leadership Computing Facility, which is a DOE Office of Science User Facility supported under Contract DEAC05-00OR22725.
This research used resources of the National Energy Research Scientific Computing Center (NERSC), a Department of Energy Office of Science User Facility using NERSC award DDR-ERCAP0030598.

This manuscript has been authored by UT-Battelle LLC under contract DE-AC05-00OR22725 with the US Department of Energy (DOE). The US government retains and the publisher, by accepting the article for publication, acknowledges that the US government retains a nonexclusive, paid-up, irrevocable, worldwide license to publish or reproduce the published form of this manuscript, or allow others to do so, for US government purposes. DOE will provide public access to these results of federally sponsored research in accordance with the DOE Public Access Plan (https://www.energy.gov/doe-public-access-plan).

\bibliography{iclr2025_conference}
\bibliographystyle{iclr2025_conference}

\newpage
\setcounter{table}{0}
\renewcommand{\thetable}{A\arabic{table}}
\setcounter{figure}{0}
\renewcommand{\thefigure}{A\arabic{figure}}

\appendix
\section{Appendix}
\subsection{Datasets}\label{app-data}

Three datasets were used in the work: PDEBench \citep{takamoto2022pdebench}, colliding thermals \citep{norman_mrnormanminiweather_2024}, and lid-driven cavity MHD flows. 
\begin{itemize}
    \item PDEBench (\url{https://github.com/pdebench/PDEBench}) consists of diverse 1D, 2D, and 3D diverse benchmark datasets. We used the 2D cases -- incompressible flows, compressible flows, turbulent flows, reaction diffusion, and shallow water -- for model pretraining in Section \ref{sec-effect-pre}. The govern equations are summarized below.
    \begin{itemize}
        \item Shallow water equations [swe]:
        \[
        \partial_t h+ \nabla\cdot (h\boldsymbol{v})=0, 
        \]
        \[
        \partial_t(h\boldsymbol{v})+\nabla\cdot\left(\frac{1}{2}h\boldsymbol{v}^2+\frac{1}{2}g_rh^2\right)=-g_rh\nabla b
        \]
        
        \item Diffusion-reaction equations [diffre2d]:
        \[
        \partial_t \boldsymbol{c} = \boldsymbol{D}\nabla^2\boldsymbol{c}+\boldsymbol{R}(\boldsymbol{c}),.
        \]
        where $\xi$ and $\phi$ in $\boldsymbol{c}=[\xi,\phi]$ are the activator and the inhibitor, respectively.
        
        \item Incompressible NS [Incomp]:
        \[
        \nabla\cdot \boldsymbol{v}=0, 
        \]
        \[
        \rho \left(\partial_t\boldsymbol{v}+\boldsymbol{v}\cdot\nabla\boldsymbol{v}\right) = -\nabla p+\eta \nabla^2\boldsymbol{v}+\boldsymbol{f}
        \]

        \item Compressible NS [compNS] with random and turbulent initial conditions:
        \[
        \partial_t\rho + \nabla\cdot (\rho\boldsymbol{v})=0, 
        \]
        \[
        \rho \left(\partial_t\boldsymbol{v}+\boldsymbol{v}\cdot\nabla\boldsymbol{v}\right) = -\nabla p+\eta \nabla^2\boldsymbol{v}+(\zeta +\eta/3)\nabla(\nabla\cdot \boldsymbol{v})
        \]
        \[
        \partial_t\left[\epsilon+\frac{\rho \boldsymbol{v}^2}{2}\right] +\nabla\cdot \left[\left(\epsilon+p+\frac{\rho \boldsymbol{v}^2}{2}\right)\boldsymbol{v}-\boldsymbol{v}\cdot\boldsymbol{\sigma}^\prime\right]=0
        \]
        with $\epsilon=p/\Gamma-1$ and $\Gamma=5/3$.

    \end{itemize}
    For more details on these cases and equations, users are referred to\citep{takamoto2022pdebench}.
    
    \item The colliding thermals dataset was generated for our work, and the details will be presented in Section \ref{app-subsec-mw}. It was used in the experiments in Sections \ref{sec-atts} and \ref{sec-adap}, and also as one of the two fine-tuning cases in Section \ref{sec-effect-pre}.
    \item Lid-driven cavity MHD dataset was also generated in our work, and it was used as the other fine-tuning case in Section \ref{sec-effect-pre}. We will present the details in Section \ref{app-subsec-mhd}.
\end{itemize}

\subsubsection{Colliding thermals}\label{app-subsec-mw}

Thermal collision datasets contains multiple time history trajectories of the mixing of two bubbles-one cold bubble at the top colliding with a warm bubble at the bottom. Details about the governing equations can be found in \cite{norman_mrnormanminiweather_2024}. 
These trajectories start from different initial temperature conditions as
\begin{equation}
    T_0(x,z) = 300.0 + T_{10}(x,z) + T_{20}(x,z),
\end{equation}
with one hot $T_{10}$ and cold $T_{20}$ thermals being
\begin{equation}
      T_{10}(x,z)=\left\{
                \begin{array}{ll}
                Tc_1\cos{\left(\frac{\pi}{2}d_1(x,z)\right)}^2, \text{ if } d_1(x,z)\leq 1\\
                  0, \text{ otherwise}
                \end{array}
              \right.
\end{equation}
and
\begin{equation}
        T_{20}(x,z)=\left\{
                \begin{array}{ll}
                - Tc_2\cos{\left(\frac{\pi}{2}d_2(x,z)\right)}^2, \text{ if }d_2(x,z)\leq 1\\
                  0, \text{ otherwise}
                \end{array}
              \right.         
\end{equation}
where $Tc_i$ is the center temperature amplitude and $d_i(x,z) = \sqrt{\frac{(x-xc_i)^2}{rx_i^2}+\frac{(z-zc_i)^2}{rz_i^2}}$ is the distance from thermal center ($xc_i$, $zc_i$) for $i=1,2$.
The thermals are elliptical in shape with the radius, $rx_i$ and $rz_i$,  in x and z directions, respectively.

\paragraph{Configurations}
We sample 4096 configurations with the thermals ($i=1,2$) at different locations following uniform distribution, 
\begin{equation}
    xc_i\sim U[0.2L,0.8L]\text{, } zc_1\sim U[0.2L, 0.3L]\text{, and } zc_2\sim U[0.7L, 0.8L],
\end{equation}
with different elliptical shapes also following uniform distribution, 
\begin{equation}
   rx_i\sim U[0.1L, 0.2L] \text{ and } rz_i \sim U[0.1L, 0.2L], 
\end{equation}
 and with temperature amplitudes equally sampled from, 
 \begin{equation}
 Tc_i\sim C\{10, 15, 20, 25\}.  
\end{equation}

The equations are solved by using a finite volume method with $nx=256, ny=256$ grid points in x and z directions, respectively. The simulations are advanced in time for 500 seconds and solutions are saved every 0.5 second. In total, we have 4096 trajectories, each with data at size ($nt=1001, nx=256, ny=256$).


\subsubsection{Lid-driven cavity magnetohydrodynamics (MHD) flows} \label{app-subsec-mhd}
The MHD dataset contains solution trajectories from initial conditions to steady states for a benchmark lid-driven cavity MHD flow problem in two dimensions with varying configurations. The MHD flow is governed by an incompressible Navier-Stokes equation with Lorentz force coupled with an induction equation with divergence cleaning. The detail formulation of the governing equations and problem setting for the lid-driven MHD cavity problem are given in \cite{FAMBRI2023112493}. 

\paragraph{Configurations} In this dataset, we include solution trajectories of the lid-driven cavity problem at three magnetic Reynolds numbers $\texttt{Re}_{\texttt{m}}=100$, $200$, and $500$, each with ten external horizontal magnetic field magnitude $B_x = 0.05, 0.10, \dots, 0.50$. This gives 30 different problem configurations.
For each problem configuration, the fluid velocity field $\mathbf{v}$ and the magnetic field $\mathbf{B}$ are recorded on a 128$\times$128 uniform spatial mesh for 2,000 time steps.

\subsection{More on spatiotemporal attentions and adaptive tokenization} \label{app-sec-algorithm}

\paragraph{Training setting} We randomly sampled a subset with 512 trajectories for training and 64 trajectories for testing for the results in Sections \ref{sec-atts} and \ref{sec-adap}.
During training, we use the \texttt{AdamW} optimizer with a learning rate equal to $10^{-4}$. Batch size was set to be 128 and accumulate gradient step was set to be 1.
Models were trained for 20,000 steps. For cases with constant patch size, the value was set to be $32\times32$.

\begin{figure}[h]
\centering
\includegraphics[width=0.6\textwidth]{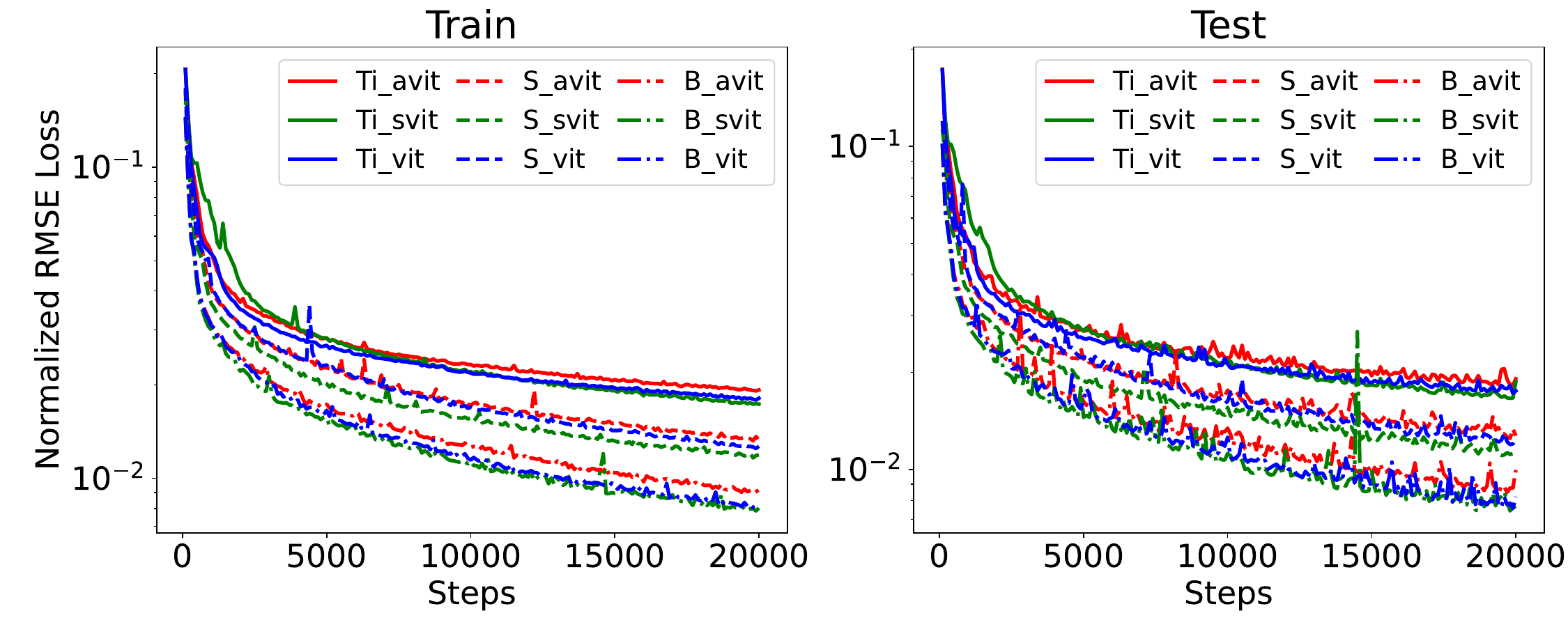}
\caption{Loss history of three spatiotemporal attention schemes at three model sizes during training}
\label{fig-att-loss}
\end{figure}

\begin{figure}[h]
\centering
\includegraphics[width=0.75\textwidth]{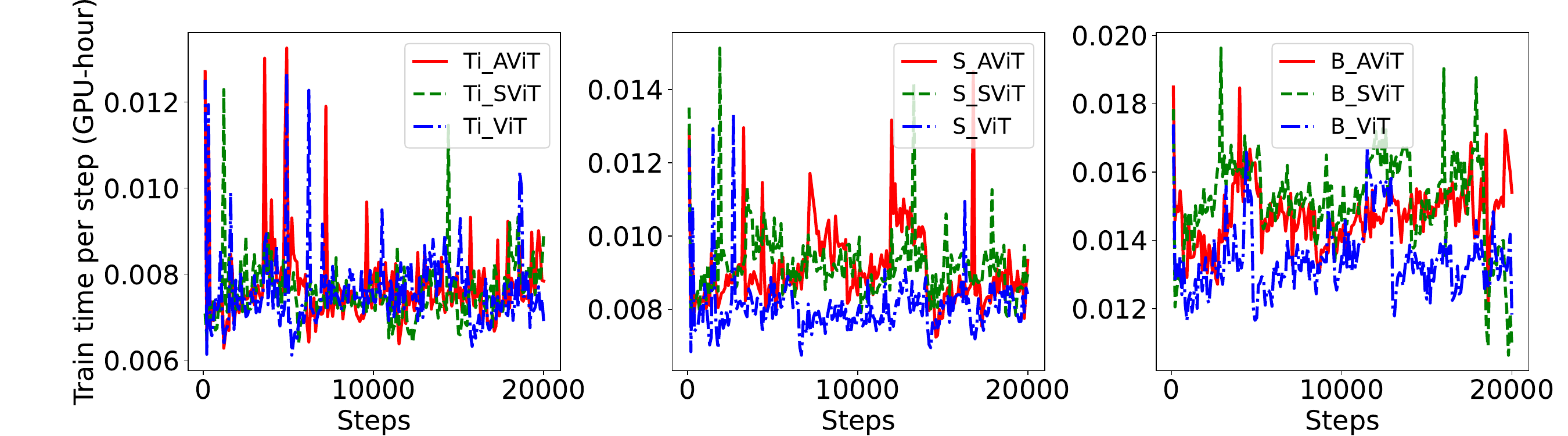}
\caption{Training time per step of three spatiotemporal attention schemes at three model sizes}
\label{fig-att-time}
\end{figure}

For the experiment on spatotemporal attention schemes in Section \ref{sec-atts}, we ran 9 cases with AViT, SViT, and ViT attention blocks at three sizes (Ti, S, and B). 
Figure~\ref{fig-att-loss} shows the loss history during training of the models for both training and test sets, and Figure~\ref{fig-att-time} shows the training time cost. 

\begin{figure}[h]
\centering
\includegraphics[width=0.45\textwidth]{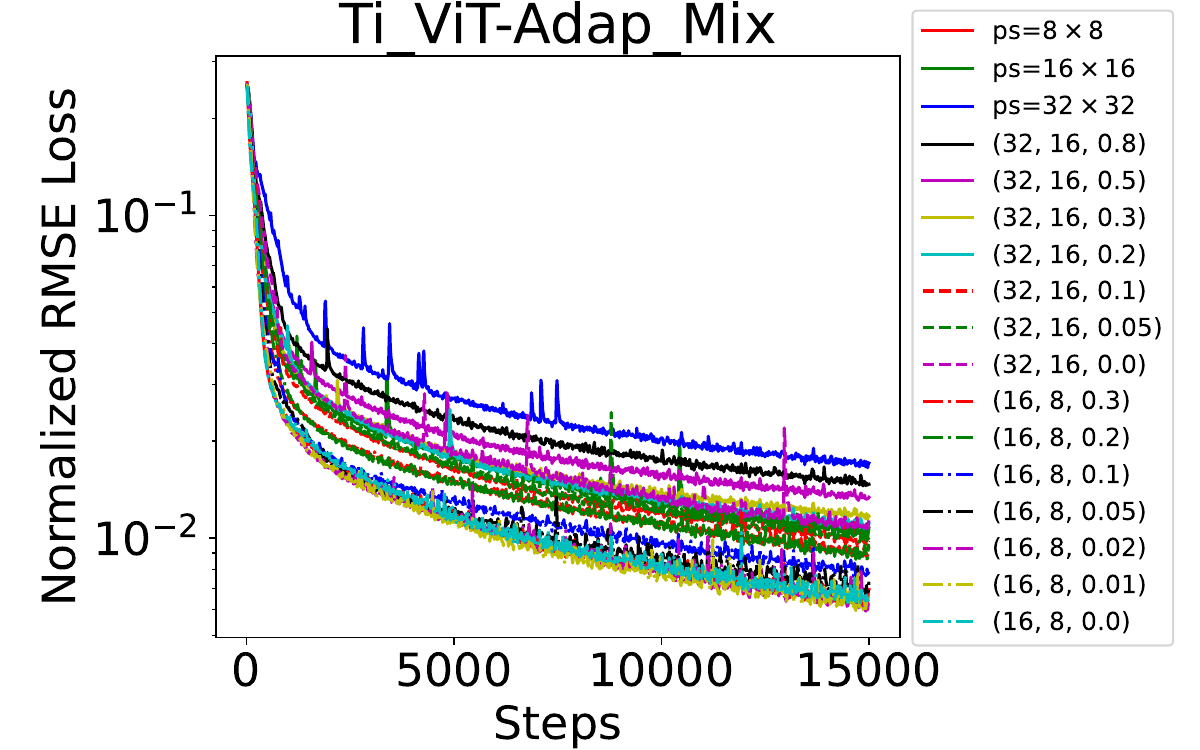}
\includegraphics[width=0.45\textwidth]{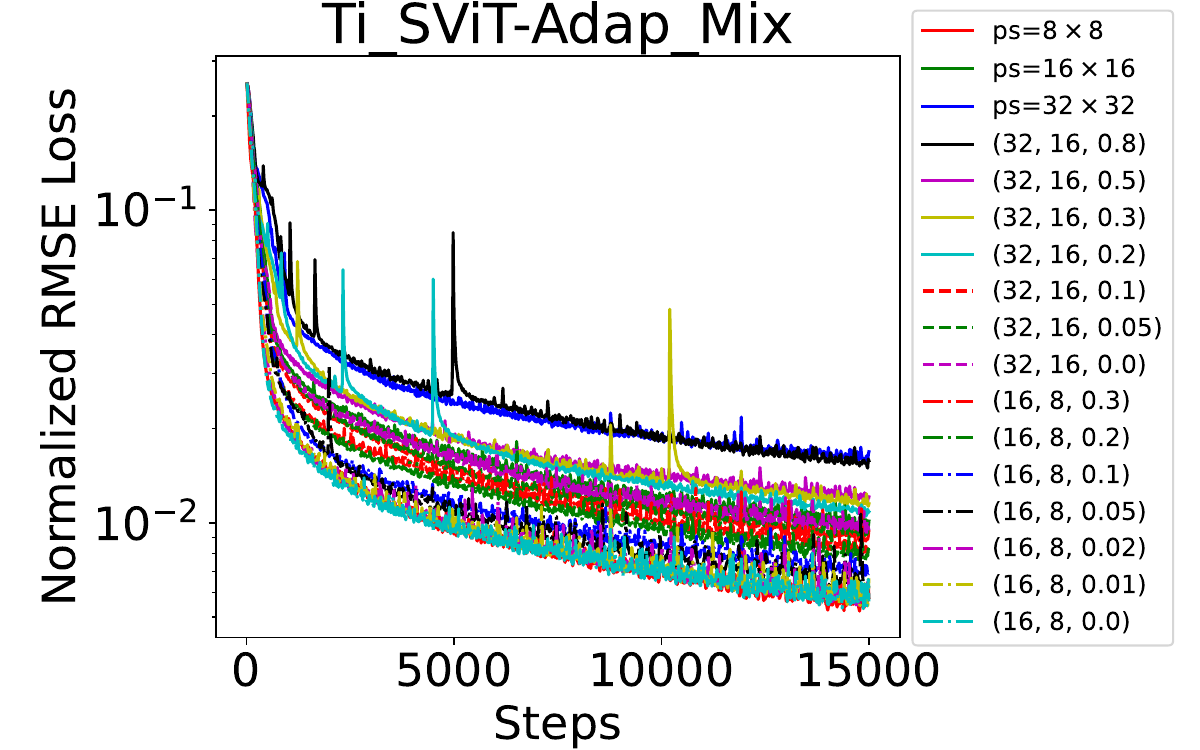}
\caption{Comparison of training loss histories of models with adaptive tokenization Adap\_Mix and constant patch sizes (ps=$32\times32$, ps=$16\times16$, and ps=$8\times8$) for the two spatiotemporal attention schemes (ViT and SViT) in a single colliding thermals trajectory.}
\label{fig-adapdummy-loss}
\end{figure}

\begin{figure}[h]
\centering
\includegraphics[width=0.6\textwidth]{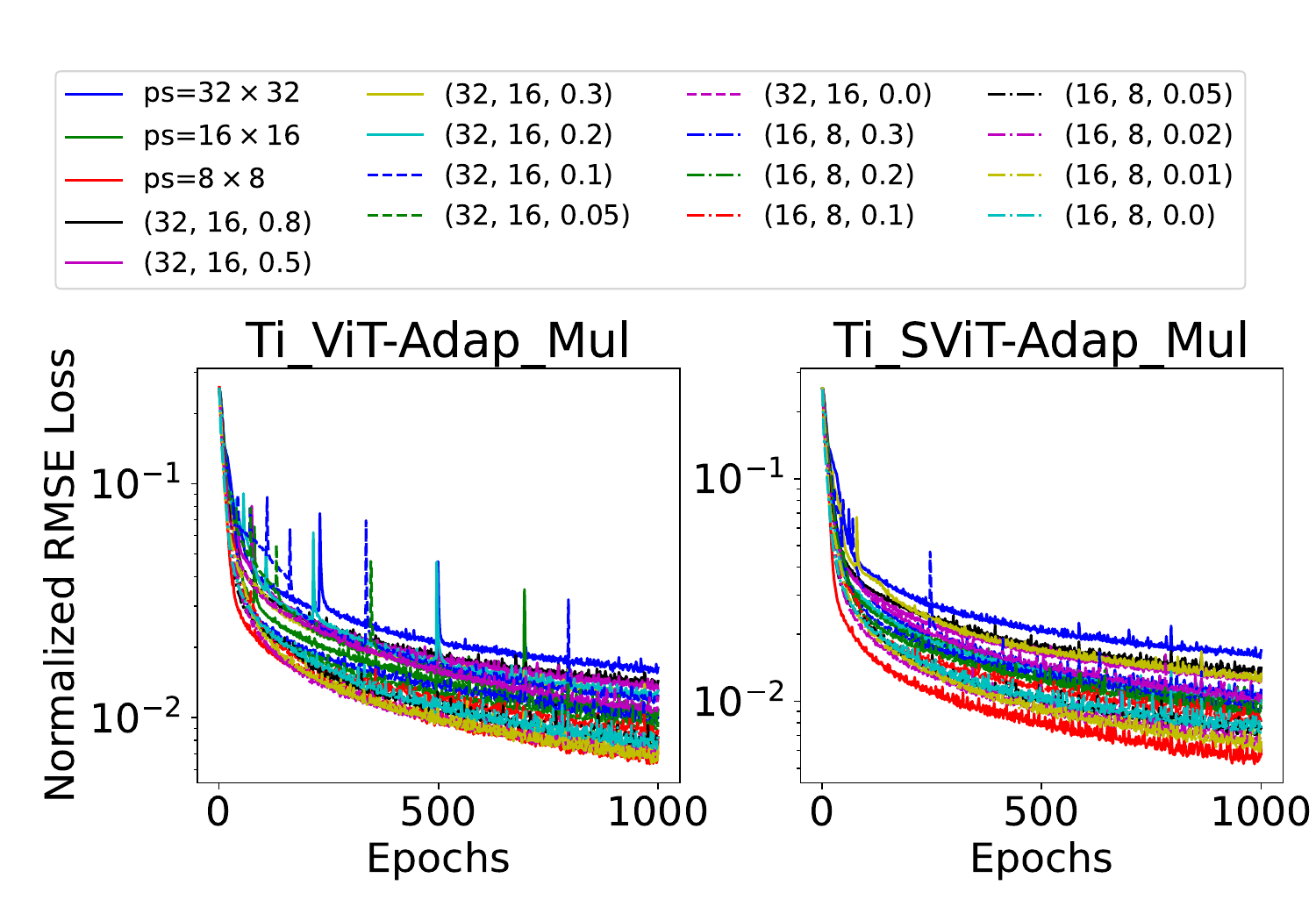}
\caption{Comparison of training loss histories of models with adaptive tokenization Adap\_Mul and constant patch sizes (ps=$32\times32$, ps=$16\times16$, and ps=$8\times8$) for the three spatiotemporal attention schemes (ViT and SViT) in a single colliding thermals trajectory.}
\label{fig-adapdummy-loss-mul}
\end{figure}

\begin{figure}[h]
\centering
\includegraphics[width=0.6\textwidth]{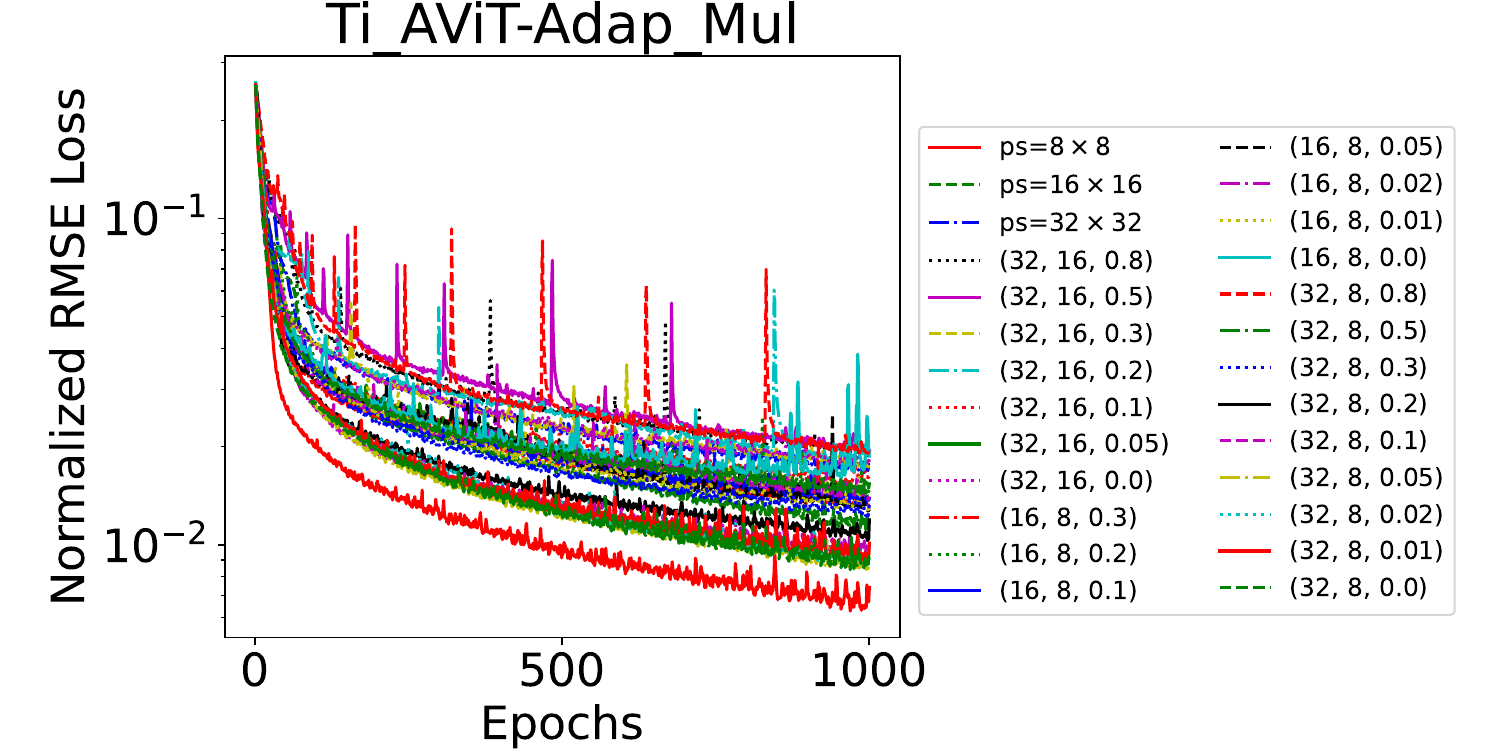}
\caption{Comparison of training loss histories of models with adaptive tokenization Adap\_Mul and constant patch sizes (ps=$32\times32$, ps=$16\times16$, and ps=$8\times8$) for AViT in a single colliding thermals trajectory.}
\label{fig-adapdummy-loss-mul-avit}
\end{figure}

For the experiment on adaptive tokenization in Section \ref{sec-adap}, Figures~\ref{fig-adapdummy-loss}, \ref{fig-adapdummy-loss-mul}, and \ref{fig-adapdummy-loss-mul-avit} show the training losses of all models in a single colliding thermal trajectory for Figures~ \ref{fig-adapdummy-gamma-adapmix}, \ref{fig-adapdummy-vit-gamma-conv}, and \ref{fig-adapdummy-vit-gamma-conv-avit}, respectively. 

\subsection{Pretraining and fine-tuning}\label{app-sec-pret-finet}

\subsubsection{Pretraining} \label{app-secsec-pre}

\begin{table}[h]
  \begin{center}
    \caption{Cases and datasets}
    \label{tab-app-data}
    \subcaption*{\textbf{Pretraining}: PDEBench \cite{takamoto2022pdebench}}
    
\begin{tabular}{l|crr} 
Dataset& Variables ($C$) & Spatiotemporal res. ($T\times H \times W$) & $N_\text{traj}$ trajectories  \\\hline
Shallow-water &  $h$ &  $101\times 128 \times 128$ & 1,000  \\
Diffusion-reaction [diffre2d] & $\xi, \phi$ & $101\times 128 \times 128$ &   1,000  \\
Incompressible NS &$u, v, \rho_\text{aug}$ & $1000\times 512 \times 512$  &  992   \\
Compressible NS Rand-128 & $u, v, \rho, P$ & $21\times 128 \times 128$  & 40,000  \\
Compressible NS Rand-512 & $u, v, \rho, P$  & $21\times 512\times 512$  & 2,000 \\
Compressible NS Turb & $u, v, \rho, P$  & $21\times 512 \times 512$  & 2,000\\
\end{tabular}
 \bigskip
\subcaption*{\textbf{Fine-tuning}: colliding thermals (Section \ref{app-subsec-mw}) and lid-driven MHD (Section \ref{app-subsec-mhd})}
\begin{tabular}{l|crr} 
Dataset& Variables ($C$) & Spatiotemporal res. ($T\times H \times W$) & $N_\text{traj}$ trajectories in training \\\hline
colliding thermals &  $\rho, u, v, T$ &  $1001\times 256 \times 256$ & [1, 6, 12, 24, 48]  \\
lid-driven MHD  & $u, v, B_x, B_y$ & $2000 \times 128 \times 128$ &   [1, 3, 6, 12, 24]  \\
 \end{tabular}
  \end{center}
\end{table}

Five 2D datasets from PDEBench \cite{takamoto2022pdebench} were used for pretraining, including shallow water, diffusion reaction, incompressible flows, compressible flows, and turbulent flows. The details of these datasets including physical variables, spatiotemporal resolutions, and number of trajectories are summarized in Table \ref{tab-app-data}. 

During training, we used the \texttt{AdamW} optimizer with \texttt{DAdaptAdam} for learning rate scheduling.
Batch size was set to be 1472 and patch size was $32\times32$.
Training/testing/validating split was 0.8/0.1/0.1.
Gradient accumulation was set to be 1.
We trained the model for 30,000 steps to predict the next step solution given a history of $T=16$.

\subsubsection{Fine-tuning} \label{app-secsec-fine}

For fine-tuning, we evaluate the transferrability of pretrained models to systems with distinct physics and different sets of variables. Table \ref{tab-app-data} summarizes the two fine-tuning cases: colliding thermals and lid-driven cavity MHD flows.
In the two cases, pretrained models were fine-tuned to predict the solution at a future time $t+t_{\textup{lead}}$ given a history of solutions from $t-T+1$ to $t$. In our experiments, $T$ was set to be 10 while $t_{\textup{lead}}$ was set to 50 for the colliding thermals and 100 for the lid-driven cavity MHD flows. The fine-tuned models were evaluated on a held-out test set for all runs in each case.
We used the \texttt{AdamW} optimizer with a learning rate equal to ${10}^{-4}$.
Batch size was set to be 256. 
Models were fine-tuned for 600 epochs for colliding thermals and 1000 epochs for lid=drive cavity MHD flows.

\paragraph{Colliding thermals} We sampled 1, 6, 12, and 24  trajectories for training. The results in Section \ref{sec-effect-pre} are shown for a fixed test set with 24 trajectories. 


\paragraph{Lid-driven cavity MHD flows} Among the 30 cases, we kept 6 for testing. From the remaining 24 cases, we sampled 1, 3, 6, and 12 cases to assess the impact of the amount of fine-tuning data.

\end{document}